\documentclass[twocolumn]{article}
\usepackage{arxiv}
\usepackage{amsmath,amssymb,amsfonts}
\usepackage[utf8]{inputenc}	
\usepackage[T1]{fontenc}	
\usepackage{xcolor}		
\usepackage[colorlinks = true,
            linkcolor = purple,
            urlcolor  = blue,
            citecolor = cyan,
            anchorcolor = black]{hyperref}	
\usepackage{booktabs} 		
\usepackage{nicefrac}		
\usepackage{microtype}		
\usepackage{lineno}		
\usepackage{float}			
\usepackage{newfloat}
\DeclareFloatingEnvironment[name={Supplementary Figure}]{suppfigure}
\usepackage{sidecap}
\sidecaptionvpos{figure}{c}
\usepackage{titlesec}
\titlespacing\section{0pt}{12pt plus 3pt minus 3pt}{1pt plus 1pt minus 1pt}
\titlespacing\subsection{0pt}{10pt plus 3pt minus 3pt}{1pt plus 1pt minus 1pt}
\titlespacing\subsubsection{0pt}{8pt plus 3pt minus 3pt}{1pt plus 1pt minus 1pt}

\usepackage{tikz,xcolor,hyperref}

\usepackage[numbers]{natbib}
\usepackage{amsmath,amssymb,amsfonts}
\usepackage{algorithmic}
\usepackage{graphicx}
\usepackage{textcomp}
\usepackage{amsthm}
\usepackage{amsmath}
\usepackage{amsfonts}
\usepackage{multirow}
\usepackage[bb=boondox]{mathalfa}
\usepackage{tikz}
\usepackage{tikz-network}
\usepackage{caption}
\usepackage{subcaption}
\usepackage{titletoc}
\usepackage{booktabs}
\usepackage{footmisc}
\usepackage{blindtext}
\usepackage{color,soulutf8}
\usepackage[utf8]{inputenc}
\usepackage[english]{babel}
\usepackage{colortbl}
\usepackage{arydshln}
\usepackage{enumitem}
\usepackage{xcolor}
\usepackage[ruled, lined, linesnumbered, longend]{algorithm2e}

\newtheorem{theorem}{Theorem}
\setlist[itemize]{align=parleft,left=0pt..1em}

\definecolor{first}{RGB}{34, 139, 34}
\definecolor{second}{RGB}{0, 101, 167}
\definecolor{third}{RGB}{187, 112, 190}

\SetKw{Break}{break}

\DeclareMathOperator*{\argmax}{\arg\!\max}
\SetKwComment{Comment}{$\triangleright$\ }{}

\makeatletter
\def\@xfootnote[#1]{%
  \protected@xdef\@thefnmark{#1}%
  \@footnotemark\@footnotetext}
\makeatother
\usetikzlibrary{shapes.geometric, arrows.meta, positioning, shadows}
\tikzset{
    base/.style = {draw, thick, fill=white, text centered, font=\sffamily},
    circnode/.style = {base, circle, minimum size=0.1cm, drop shadow},
    rectnode/.style = {base, rectangle, minimum size=0.1cm, rounded corners, drop shadow},
    arrow/.style = {thick,-Stealth},
    line/.style = {thick}
}
\newcommand{\bigboxplus}{\mathop{\vcenter{\hbox{\tikz{
            \draw[line width=0.5pt, scale=2.25] (0,0) rectangle (1ex,1ex);
            \draw[line width=0.5pt, scale=2.25] (0.5ex,0) -- (0.5ex,1ex);
            \draw[line width=0.5pt, scale=2.25] (0,0.5ex) -- (1ex,0.5ex);
}}}}\displaylimits}
\setlist[itemize]{align=parleft,left=0pt..1em}
\usepackage{listings}
\lstdefinestyle{mystyle}{
    language=Python,
    basicstyle=\ttfamily\scriptsize,
    commentstyle=\color[rgb]{0.13,0.55,0.13},
    keywordstyle=\color[rgb]{0,0,1},
    numberstyle=\color[rgb]{0.5,0.5,0.5},
    stringstyle=\color[rgb]{0.7,0.2,0.4},
    breakatwhitespace=false,         
    breaklines=true,                 
    captionpos=b,                    
    keepspaces=true,                 
    showspaces=false,                
    showstringspaces=false,
    showtabs=false,                  
    tabsize=1,
    lineskip=-1pt
}
\newcommand{\suchthat}{\;\ifnum\currentgrouptype=16 \middle\fi|\;}
\newcommand*{\horzbar}{\rule[.5ex]{2.5ex}{0.5pt}}

\newcommand{\graphRandomExample}{%
  \resizebox{1.5cm}{!}{%
    \begin{tikzpicture}[>=stealth',rotate=90]
      \Vertex[color=gray!60, x=0, y=0, size=0.3]{1}
      \Vertex[color=gray!60, x=-0.2, y=-0.7, size=0.3]{3}
      \Vertex[color=gray!60, x=-0.1, y=0.5, size=0.3]{4}
      \Vertex[color=gray!60, x=-0.6, y=-0.3, size=0.3]{6}
      \Vertex[color=gray!60, x=-0.8, y=-1.7, size=0.3]{9}
      \Vertex[color=gray!60, x=-0.7, y=-1.2, size=0.3]{12}
      \Vertex[color=gray!60, x=0.1, y=-1.4, size=0.3]{13}
      \Vertex[color=gray!60, x=0, y=-2.0, size=0.3]{15}
      \Edge[color=black](1)(3) \Edge[color=black](1)(4) \Edge[color=black](1)(6)
      \Edge[color=black](4)(6) \Edge[color=black](6)(12) \Edge[color=black](3)(13)
      \Edge[color=black](3)(9) \Edge[color=black](3)(6) \Edge[color=black](12)(9)
      \Edge[color=black](12)(13) \Edge[color=black](9)(15) \Edge[color=black](13)(15)
    \end{tikzpicture}%
  }%
}

\newcommand{\rowMajorMatrix}{
  \resizebox{2.1cm}{!}{
    $\left[
      \begin{array}{ccc}
        \horzbar & x^\top_{0} & \horzbar  \\
                 & \vdots     &           \\
        \horzbar & x^\top_{7} & \horzbar  \\  
      \end{array}
    \right]$
  }
}

\newcommand{\thetaZeroLearnable}{
  \resizebox{2cm}{!}{
    $\left[
      \begin{array}{cccc}
        \theta_{00} & \theta_{01} & \cdots & \theta_{0d} \\
        \theta_{10} & \theta_{11} & \cdots & \theta_{1d} \\
        \vdots & \vdots & \ddots & \vdots \\
        \theta_{f0} & \theta_{f1} & \cdots & \theta_{fd}
      \end{array}
    \right]$
  }
}

\newcolumntype{C}{>{\centering\arraybackslash}p{2.2cm}}
\usetikzlibrary{matrix, positioning}
\newcommand\blfootnote[1]{%
  \begingroup
  \renewcommand\thefootnote{}\footnote{#1}%
  \addtocounter{footnote}{-1}%
  \endgroup
}

\title{Efficient Mixed Precision Quantization in Graph Neural Networks}

\usepackage{authblk}

\author[1,2]{Samir Moustafa $^\ast$}
\author[1,3]{Nils Kriege}
\author[1]{Wilfried N. Gansterer}

\affil[1]{Faculty of Computer Science, University of Vienna, Vienna, Austria}
\affil[2]{UniVie Doctoral School Computer Science, University of Vienna, Vienna, Austria}
\affil[3]{Research Network Data Science, University of Vienna, Vienna, Austria}
\affil[$\ast$]{Corresponding author}

\begin{document}

\twocolumn[ 
  \begin{@twocolumnfalse} 

\maketitle

\begin{abstract}
    Graph Neural Networks (GNNs) have become essential for handling large-scale graph applications. However, the computational demands of GNNs necessitate the development of efficient methods to accelerate inference. Mixed precision quantization emerges as a promising solution to enhance the efficiency of GNN architectures without compromising prediction performance. Compared to conventional deep learning architectures, GNN layers contain a wider set of components that can be quantized, including message passing functions, aggregation functions, update functions, the inputs, learnable parameters, and outputs of these functions.
    In this paper, we introduce a theorem for efficient quantized message passing  to aggregate integer messages. It guarantees numerical equality of the aggregated messages using integer values with respect to those obtained with full (FP32) precision. Based on this theorem, we introduce the Mixed Precision Quantization for GNN (MixQ-GNN) framework, which flexibly selects effective integer bit-widths for all components within GNN layers.
    Our approach systematically navigates the wide set of possible bit-width combinations, addressing the challenge of optimizing efficiency while aiming at maintaining comparable prediction performance. 
    MixQ-GNN integrates with existing GNN quantization methods, utilizing their graph structure advantages to achieve higher prediction performance. On average, MixQ-GNN achieved reductions in bit operations of $5.5 \times$ for node classification and $5.1 \times$ for graph classification compared to architectures represented in FP32 precision.
\end{abstract}
\vspace{0.35cm}

  \end{@twocolumnfalse} 
] 

\section{Introduction}\blfootnote{
    
    \footnotesize{Published in: 2025 IEEE 41st International Conference on Data Engineering (ICDE). DOI bookmark: \href{https://www.computer.org/csdl/proceedings-article/icde/2025/360300e038/26FZCgBUg4U}{10.1109/ICDE65448.2025.00301}.
    
    \copyright 2025 IEEE. Personal use of this material is permitted. Permission from IEEE must be obtained for all other uses, in any current or future media, including reprinting/republishing this material for advertising or promotional purposes, creating new collective works, for resale or redistribution to servers or lists, or reuse of any copyrighted component of this work in other works.}}
    As deep learning continues to gain popularity as an effective tool for embedding intelligence in electronic devices~\cite{warden2019tinyml, lin2022ondevice, Augustin2016ASO, Nagel2021AWP}, there is an increasing demand for compact, low latency, and energy-efficient models. Nowadays, GNNs are integrated into a wide range of applications that require low inference time~\cite{Zhiqiang2024LLGNN, Austin2021ETAPrediction, Weijing2020PointGNN}. However, GNN architectures can be very costly computationally due to the influence of the graph size. In comparison to deep learning architectures, GNNs have few parameters but require many computations, usually more than deep learning models~\cite{tailor2021degreequant}.
    
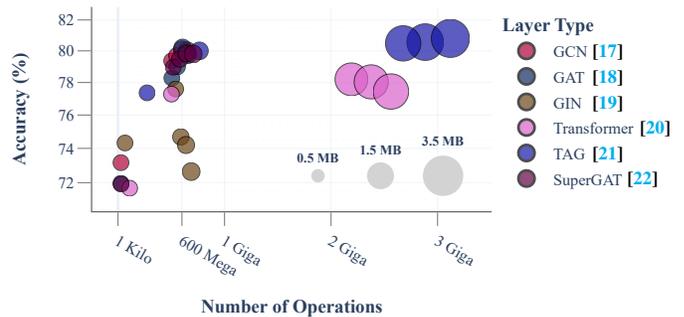
\begin{figure}
        \centering
        \begin{tikzpicture}
            \node[anchor=south west, inner sep=0] (image) at (0,0) {\includegraphics[width=0.95\linewidth]{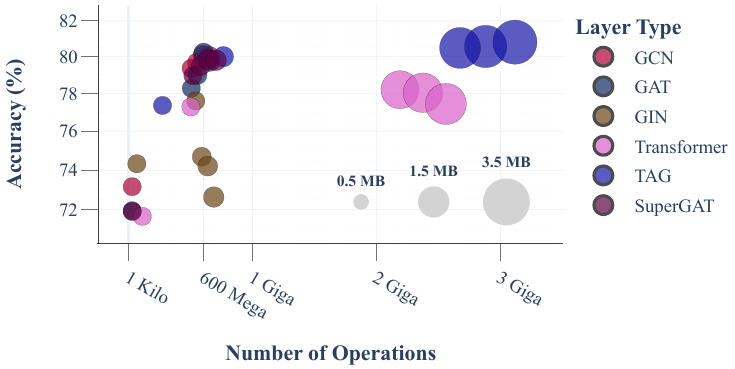}};
            \begin{scope}[x={(image.south east)},y={(image.north west)}]
                \node[black, font=\bfseries] at (0.95, 0.84) {\scriptsize \cite{kipf2017semisupervised}};
                \node[black, font=\bfseries] at (0.95, 0.76) {\scriptsize \cite{velickovic2018graph}};
                \node[black, font=\bfseries] at (0.95, 0.68) {\scriptsize \cite{xu2018how}};
                \node[black, font=\bfseries] at (1.025, 0.60) {\scriptsize \cite{ijcai2021p214}};
                \node[black, font=\bfseries] at (0.95, 0.52) {\scriptsize \cite{Du2017TopologyAG}};
                \node[black, font=\bfseries] at (1.005, 0.44) {\scriptsize \cite{kim2021how}};
            \end{scope}
        \end{tikzpicture}
        \caption{Accuracies of six types of GNNs on the Cora dataset over the number of operations required for a single forward pass (inference). Each architecture uses 1 to 5 layers of the same type, and each was tested five times with the best hyper-parameter settings. Colors distinguish the layer types, and the radius of the circles indicates the model size.}
        \label{figure:core_ops_vs_accuracy}
    \end{figure}

    In Figure~\ref{figure:core_ops_vs_accuracy}, the number of mathematical operations (OPs) between scalar values required by different GNN architectures on the $x$-axis, and the accuracy achieved by each architectures on the Cora dataset on the $y$-axis. As indicated, a correlation between the number of operations and the accuracy gained is observed, with a Spearman's rank correlation coefficient of $0.64$ and p-value $1.6 \times 10^{-4}$. The low p-value suggests a low probability of observing this correlation by chance, indicating a significant relationship between the number of operations and accuracy.
    In contrast to deep neural networks~{\cite{Goodfellow-et-al-2016}}, deeper GNN architectures often demonstrate similar or lower accuracy than smaller architectures due to over-smoothing~{\cite{Chen2019MeasuringAR}}, over-squashing~{\cite{topping2022understanding}}, structural mismatches~{\cite{Peng2024BeyondOU}}, optimization challenges~{\cite{Xu2021OptimizationOG}}, and data sparsity. Smaller models, on the other hand, achieve a balance between capacity and practicality~{\cite{Wu2022GraphNN}}.
    
    Quantization reduces the computational cost per operation and accelerates inference without changing the architecture~\cite{Wentao2021EdgeComputing}. Recent quantization methods for GNNs highlight that the main quantization error is due to quantizing high in-degree nodes with low precision~\cite{tailor2021degreequant, zhu2023rm}. Each of these works introduces different ways to use mixed precision within the definition of the Message Passing Neural Network (MPNN) to assign high precision to high in-degree nodes during message passing.
    In this work, We consider a mixed precision approach across message passing, aggregation, and update functions, as well as the inputs, learnable parameters, and outputs of these functions. We use the term \emph{component} to denote each of these elements, thereby covering all integral parts of the GNNs. For each component, the quantization is fixed to a specific bit-width. This restriction arises from the hardware limitation that mixed precision within the same component (matrix or result from matrix operations) is not applicable, or requires additional computations~\cite{andersch2022nvidia, ladder-osdi24, nvidia_blackwell_architecture}.
    For example, to perform sparse-dense matrix multiplication between a sparse matrix (adjacency matrix) and a dense matrix (node features), the numerical precision of both non-zero values of the sparse matrix, and elements of the dense matrix must be identical, or one of them must be cast to the precision of the other~{\cite{Ozaki2024ExtensionOA}}.
    
    In contrast to related work, we use mixed precision across GNN components instead of using mixed precision across the graph's nodes~\cite{tailor2021degreequant, zhu2023rm}.
    For example, in two-layer Graph Convolutional Networks (GCN)~\cite{kipf2017semisupervised}, there are nine components to be quantized: the inputs are $X$, $\hat{A}^{(1)}$, and $\hat{A}^{(2)}$. The learnable parameters for the first and second layers are $\Theta^{(1)}$ and $\Theta^{(2)}$, respectively. The outputs of each function are $X \Theta^{(1)}$, $\hat{A}^{(1)} X \Theta^{(1)}$, $\sigma(\hat{A}^{(1)} X \Theta^{(1)}) \Theta^{(2)}$, where $\sigma$ is the ReLU activation function, and prediction output $\hat{A}^{(2)} \sigma(\hat{A}^{(1)} X \Theta^{(1)}) \Theta^{(2)}$.
    
    \begin{figure}
        \centering
        \includegraphics[width=0.95\linewidth]{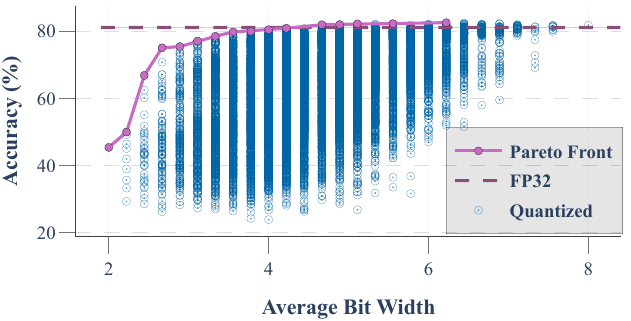}
        \caption{Accuracy vs. average bit-width for a two layer GCN architecture on the Cora dataset, using mixed precision quantization across 9 components within the architecture. Bit-width options: $\{2, 4, 8\}$, totaling $3^9$ possible combinations, each point \textcolor[HTML]{0065A7}{$\odot$} is an instance of this combination.}
        \label{figure:all_options_for_gcn_over_cora}
    \end{figure}
    
    \begin{figure*}
        \centering
        \includegraphics[width=\linewidth]{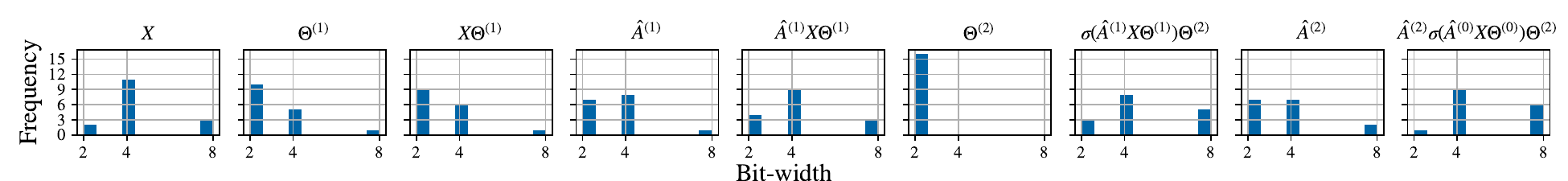}
        \caption{Histograms of the bit-widths for each component in the two layer GCN architecture on the Pareto front of Figure~\ref{figure:all_options_for_gcn_over_cora}. The 16 different choices along the Pareto front line are considered the optimal bit-width combinations.}
        \label{figure:pareto_front_distribution}
    \end{figure*}
    
    In Figure~\ref{figure:all_options_for_gcn_over_cora}, the accuracies achieved by the two layer GCN architecture on the Cora dataset under different precision levels are shown. The points represent all possible combinations of different bit-widths $\{2, 4, 8\}$ over the nine components in the two-layer GCN architecture, with each point representing the mean of three runs of the quantized architecture. Since the architecture, dataset, and quantization method are the same for each point, the average bit-width is a proxy to measure the efficiency of each point. In total, $19,683$ combinations are considered. This makes exhaustive exploration of the search space computationally infeasible for large-scale datasets or deeper architectures. As shown, a set of quantized candidates statistically outperform the full precision (FP32) version. To clearly demonstrate the complexity of selecting optimal bit-widths for the components of a simple 2-layer GCN on the Cora dataset, Figure~\ref{figure:all_options_for_gcn_over_cora} shows all the available options. Subsequently, Figure~\ref{figure:pareto_front_distribution} illustrates the distribution of bit-widths across each component for candidates on the Pareto front. This visualization underscores the non-trivial nature of the distribution, particularly highlighting the significant variation in bit-widths, which do not follow a specific pattern. This illustrates that identifying the most effective bit-width (quantization) configuration for enhancing prediction performance is challenging.
    
    \textbf{Main contributions:}
        1) We introduce a theorem which is the basis for an efficient quantized message passing framework using integer representations. This method employs pre-computed factors, ensuring that integer-based message aggregation yields results equivalent to those of the original full-precision approach.
        
        2) We introduce the Mixed precision Quantization for GNN (MixQ-GNN) framework which is designed to efficiently search for highly effective combinations of bit-width choices. It tackles the challenge posed by the extensive number of potential bit-width combinations across different components in GNNs. Our MixQ-GNN approach can also be efficiently integrated with existing message passing quantization schemes~\cite{tailor2021degreequant}.
        
\section{Preliminaries}    
    \textbf{Graph notation:} 
    A graph $\mathcal{G}$ is a tuple $(V, E, X, W)$, where $V = \{v_1, \cdots, v_n\}$ is the set of $n$ nodes, $E \subseteq V \times V$ is the set of $m$ edges, $X \in \mathbb{R}^{n \times f}$ is the node feature matrix ($X_{i,:} = x_{v_i}$) such that each node has $f$ features, and $W$ represents edge weights, where each $w_{ij} \in W$ corresponds to the weight of the edge $e_{ij}$ from node $v_i$ to node $v_j$. The adjacency matrix $A \in \mathbb{R}^{n \times n}$ of $\mathcal{G}$ is a sparse matrix, where $A_{i,j}=w_{ij}$ if an edge $e_{ij} \in E$ exists, and $0$ otherwise.

    \textbf{Graph Neural Network} layers perform message passing, aggregation, and update functions, forming the MPNN framework~\cite{Gilmer2017MPNN}. Messages from neighbors are aggregated and combined with embeddings to update each node, as in Equation~(\ref{equation:message_passing_neural_networks}), where $\mathcal{N}^{i}(v)$ denotes the set of in-neighbors of node $v$, which is used to construct the depth-1 unfolding tree for node $v$. $\bigoplus$ is a permutation invariant function, and $\mathcal{U}^{(l)}, \mathcal{M}^{(l)}$ are transformations at layer $l$, such $\mathcal{U}^{(l)}, \mathcal{M}^{(l)} : \mathbb{R}^{M \times N} \rightarrow \mathbb{R}^{M \times K}$ for all $M, N, K \geq 1$. Initial embedding for node $v$ starts with $h^{(0)}_v = x_{v}$, and the $l^\text{th}$ layer embedding: 
    \begin{equation}
        h_{v}^{(l)}=
        \overbrace{\mathcal{U}^{(l)}\Bigl[\underbrace{\bigoplus_{u \in \mathcal{N}^i(v)} \Bigl[ \overbrace{\mathcal{M}^{(l)}\Bigl[h_{u}^{(l-1)}\Bigr]}^{\text{Message}}, w_{vu}\Bigr]}_{\text{Aggregate}}\Bigr]}^{\text{Update}}.
        \label{equation:message_passing_neural_networks}
    \end{equation}
    The matrix formulation in Equation~(\ref{equation:message_passing_neural_networks_matrix}) uses the adjacency matrix $A$ to aggregate messages across the graph, demonstrating the parallel update capabilities of GNNs. This formulation is equivalent to the MPNN in Equation~(\ref{equation:message_passing_neural_networks}). $H^{(l)}$ represents embedding matrix of all nodes at layer $l$ where $H^{(l)}_{i,:} = h^{(l)}_v$, and $\bigboxplus$ represents aggregation operator equivalent to $\bigoplus$. For simplicity, and without losing generality, we can set $\bigboxplus$ as the matrix multiplication operator, and the equivalent form for it would be to set $\bigoplus$ as sum function $\sum$.
    \begin{equation}
        H^{(l)} = \overbrace{\mathcal{U}^{(l)}\Bigl[\underbrace{\bigboxplus \Bigl[ A,\; \overbrace{\mathcal{M}^{(l)}\Bigl[H^{(l-1)}\Bigr]}^{\text{Message}} \Bigr]}_{\text{Aggregate}}\Bigr]}^{\text{Update}}
        \label{equation:message_passing_neural_networks_matrix}
    \end{equation}
    There is a considerable number of GNN architectures that can be formulated using message-passing schema and learnable non-linear transformations~\cite{kipf2017semisupervised, xu2018how, velickovic2018graph, Hamilton2017InductiveRL, ijcai2021p214, Du2017TopologyAG}.
    Adjustments to edge set $E$, or edge weight set $W$, correspond to modifications in the adjacency matrix $A$.
    
    In Graph Convolutional Networks (GCN)~\cite{kipf2017semisupervised}, the edge weight $\hat{w}_{ij}$ is defined as $\hat{w}_{ij} = w_{ij} / \sqrt{d_i d_j}$, where $d_v = 1 + \sum_{u \in \mathcal{N}^{i}(v)} w_{vu}$. This corresponds to the adjacency matrix $\hat{A}$, defined as $\hat{A} = D^{-\frac{1}{2}} (I + A) D^{-\frac{1}{2}}$, where $D$ is a diagonal matrix, with $D_{i,i} = (\sum_{j=0} (I + A)_{:,j})_{i,:}$. Setting $\mathcal{M}^{(l)}$ to be a learnable linear transformation with parameter $\Theta^{(l)}$, and $\mathcal{U}^{(l)}$ to the non-linear function $\sigma$, the definition of the GCN becomes $h_{v}^{(l)} = \sigma \left(\left(\sum_{u \in \mathcal{N}^i(v)} h_{u}^{(l-1)} \hat{w}_{vu}\right) \Theta^{(l)}\right)$, or in matrix form, $H^{(l)} = \sigma \left(\hat{A} H^{(l-1)} \Theta^{(l)} \right)$.

    In Graph Isomorphism Networks (GIN)~\cite{xu2018how}, the graph is unweighted ($w_{vu} = 1$), $\mathcal{M}^{(l)}$ is set to be the identity function, and $\mathcal{U}^{(l)}$ contains a scalar learnable parameter $\epsilon$ where $(1 + \epsilon)$ multiplies by the node feature(s) and adds the result to the aggregated messages, followed by a multi-layer perceptron (MLP) which is a stack of learnable non-linear transformations. In mathematical notation, the definition of GIN can be written as $h_{v}^{(l)} = \text{MLP}\left((1 + \epsilon) \; h_{v}^{(l-1)} + \sum_{u \in \mathcal{N}^i(v)} h_{u}^{(l-1)}\right)$, or in matrix form, $H^{(l)} = \text{MLP}\left( (1 + \epsilon) H^{(l-1)} + A H^{(l-1)} \right)$.

    In GraphSAGE~\cite{Hamilton2017InductiveRL}, node embeddings are updated by aggregating neighbor embeddings and combining them with the root node's transformed embeddings. For each node, neighbor embeddings are aggregated and transformed using a learnable weight matrix $\Theta_2$, capturing structural and feature relationships. The root node's features are independently transformed using another weight matrix $\Theta_1$, and this is the $\mathcal{M}^{(l)}$ function and the update function $\mathcal{U}^{(l)}$ to the non-linear function $\sigma$. The final embeddings are computed as $h_{v}^{(l)} = \sigma\Bigl(\Theta^{(l)}_1 h_{v}^{(l - 1)} + \Theta^{(l)}_2 \sum_{u \in \mathcal{N}(v)} h_{u}^{(l-1)}\Bigr)$, or in matrix form, $H^{(l)} = \sigma\Bigl(\Theta_1^{(l)} H^{(l-1)} + \Theta_2^{(l)} (A H^{(l-1)})\Bigr)$.

    \textbf{Quantization} of a neural network maps a real numerical parameter to an integer representation, denoted as $Q(\cdot): \mathbb{R} \rightarrow \mathbb{Z}$. Directly quantizing a pre-trained model with real parameters to a low precision integer representation, however, typically reduces prediction quality drastically due to the induction of large rounding errors~\cite{Bengio2013EstimatingOP}. Therefore, fake quantization (simulated quantization) is used during training, expressed as $Q^f = Q^{-1}(Q(\cdot)): \mathbb{R} \rightarrow \mathbb{Z} \rightarrow \mathbb{R}$, and is removed during inference, where operations are performed using discrete representations~\cite{Benoit2018Quantization, Nagel2021AWP}. We consider quantization-aware training (QAT), where the quantization function $Q$ and the de-quantization function $Q^{-1}$ are parameterized by a scale vector $S$ and a zero-point integer vector (offset) $Z$, which are tuned during training via gradient-based optimization. The quantization and de-quantization functions are defined by equations (\ref{equation:quantization_function}) for quantization and (\ref{equation:de_quantization_function}) for de-quantization.
    \begin{align}
        Q(X) &= \text{clip}\left(\lfloor{X \oslash S}\rceil + Z, a, b \right) \label{equation:quantization_function} \\
        Q^{-1}(X) &= \left(X - Z\right) \odot S \label{equation:de_quantization_function}
    \end{align}
    $\lfloor . \rceil$ denotes the nearest integer rounding function, $\text{clip}(x, a, b) = \max(a, \min(x, b))$ is a clipping function within the defined range $[a, b]$, and $\oslash$ and $\odot$ denote element-wise division and element-wise multiplication, respectively.
    
    Due to the use of non-differentiable nearest integer rounding $\lfloor . \rceil$, a surrogate method for the gradient flow can be utilized by defining a custom gradient backward pass using Straight-Through Estimators (STE)~\cite{Bengio2013EstimatingOP} to apply the backpropagation by treating non-differentiable intervals as identity functions.

\section{Related Work}
    In recent years, efficiency of GNN inference has emerged as an important topic due to the substantial computational cost. Several studies have aimed to reduce the runtime or the computational demands of GNN architectures during inference time. These studies can be categorized into hardware-based studies~\cite{Yidi2021VertexCentric, Iqbal2021ReGraphX, Zhangxiaowen2022Graphite, Zhiqiang2022Graphiler, Xiaobing2022Rubik} and algorithmic-methodological studies. The latter primarily focus on reducing computational cost, as there is an inevitable physical limit to hardware enhancements or scale. 
    Algorithmic-methodological approaches can be grouped into: (1)~Input graph or feature modification for a more efficient representation, which includes sparsification~\cite{Chen2021AUL, Liu2022ComprehensiveGG}, sampling~\cite{Zeng2020GraphSAINT, Fey2021GNNAutoScaleSA}, and vector quantization~\cite{Ding2021VQGNNAU, Huang2022EPQuantAG, Feng2020SGQuantST}. (2)~Architectural design for creating efficient architectures, involving neural architecture search for GNNs~\cite{Gao2020GraphNA, Gao2023GraphNASDA}, and knowledge distillation~\cite{Yang2020DistillingKF, zhang2021graphless}. (3)~Binarized and quantized GNNs designed to reduce the computational cost. Both (1) and (2) aim to enhance efficiency by modifying either the input or the architecture without considering that, for GNNs, the expressivity of the architectures is dependent on the input graph~\cite{aamand2022exponentially}.
    
    In this paper, we focus on approaches from group~(3) which improve computational efficiency while maintaining prediction performance. Generally, the binarization of GNNs leads to a significant loss in prediction performance due to constraints within the binary space~\cite{Jing2021MetaAggregatorLT, Wang2020BinarizedGN, Bahri2020BinaryGN}. Quantization, on the other hand, reduces the bit-width of components in GNNs to lower precision by replacing floating-point (FP) computations with integer (INT) computations~\cite{tailor2021degreequant, zhu2023rm}. Overall, the quantization of GNNs demonstrates a promising direction for improving computational efficiency, without compromising either expressivity~\cite{aamand2022exponentially} or prediction performance~\cite{tailor2021degreequant, zhu2023rm}. 

    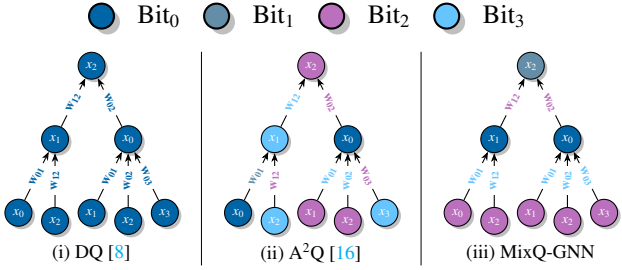
\begin{figure}
        \centering
        \definecolor{fp32_color}{RGB}{167, 42, 76}
\definecolor{int8_color}{RGB}{0, 101, 167}
\definecolor{int4_color}{RGB}{100, 141, 167}
\definecolor{int2_color}{RGB}{187, 112, 190}
\definecolor{int1_color}{RGB}{102, 196, 255}

\newcommand{\Legend}{
    \begin{tikzpicture}[node distance=0em and 0.5em]
        \node[circnode, fill=int8_color] (l2) {};
        \node[right=of l2] (label2) {Bit$_0$};
        \node[circnode, fill=int4_color, right=of label2] (l3) {};
        \node[right=of l3] (label3) {Bit$_1$};
        \node[circnode, fill=int2_color, right=of label3] (l4) {};
        \node[right=of l4] (label4) {Bit$_2$};
        \node[circnode, fill=int1_color, right=of label4] (l5) {};
        \node[right=of l5] (label5) {Bit$_3$};
    \end{tikzpicture}
}

\newcommand{\DQExample}{
    \begin{tikzpicture}[node distance=4em and 1.25em]
      \node[circnode, fill=int8_color, text=white] (x2) {$x_2$};
      \node[circnode, fill=int8_color, below left = of x2, text=white] (x1) {$x_1$};
      \node[circnode, fill=int8_color, below right = of x2, text=white] (x0) {$x_0$};
      \node[circnode, fill=int8_color, below left = of x1, text=white] (x10) {$x_0$};
      \node[circnode, fill=int8_color, below = of x1, text=white] (x12) {$x_2$};
      \node[circnode, fill=int8_color, below left = of x0, text=white] (x01) {$x_1$};
      \node[circnode, fill=int8_color, below = of x0, text=white] (x02) {$x_2$};
      \node[circnode, fill=int8_color, below right = of x0, text=white] (x03) {$x_3$};
    
      \draw[arrow] (x1) -- (x2) node[midway, text=int8_color, fill=white, sloped] {$\mathbf{w_{12}}$};
      \draw[arrow] (x0) -- (x2) node[midway, text=int8_color, fill=white, sloped] {$\mathbf{w_{02}}$};
      \draw[arrow] (x10) -- (x1) node[midway, text=int8_color, fill=white, sloped] {$\mathbf{w_{01}}$};
      \draw[arrow] (x12) -- (x1) node[midway, text=int8_color, fill=white, sloped] {$\mathbf{w_{12}}$};
      \draw[arrow] (x01) -- (x0) node[midway, text=int8_color, fill=white, sloped] {$\mathbf{w_{01}}$};
      \draw[arrow] (x02) -- (x0) node[midway, text=int8_color, fill=white, sloped] {$\mathbf{w_{02}}$};
      \draw[arrow] (x03) -- (x0) node[midway, text=int8_color, fill=white, sloped] {$\mathbf{w_{03}}$};
    \end{tikzpicture}
}

\newcommand{\AQExample}{
    \begin{tikzpicture}[node distance=4em and 1.25em]
          \node[circnode, fill=int2_color, text=white] (x2) {$x_2$};
          \node[circnode, fill=int1_color, text=white, below left = of x2] (x1) {$x_1$};
          \node[circnode, fill=int8_color, text=white, below right = of x2] (x0) {$x_0$};
          \node[circnode, fill=int8_color, text=white, below left = of x1] (x10) {$x_0$};
          \node[circnode, fill=int1_color, text=white, below = of x1] (x12) {$x_2$};
          \node[circnode, fill=int2_color, text=white, below left = of x0] (x01) {$x_1$};
          \node[circnode, fill=int2_color, text=white, below = of x0] (x02) {$x_2$};
          \node[circnode, fill=int1_color, text=white, below right = of x0] (x03) {$x_3$};
        
          \draw[arrow] (x1) -- (x2) node[midway, fill=white, text=int1_color, sloped] {$\mathbf{w_{12}}$};
          \draw[arrow] (x0) -- (x2) node[midway, fill=white, text=int2_color, sloped] {$\mathbf{w_{02}}$};
          \draw[arrow] (x10) -- (x1) node[midway, fill=white, text=int4_color, sloped] {$\mathbf{w_{01}}$};
          \draw[arrow] (x12) -- (x1) node[midway, fill=white, text=int2_color, sloped] {$\mathbf{w_{12}}$};
          \draw[arrow] (x01) -- (x0) node[midway, fill=white, text=int1_color, sloped] {$\mathbf{w_{01}}$};
          \draw[arrow] (x02) -- (x0) node[midway, fill=white, text=int1_color, sloped] {$\mathbf{w_{02}}$};
          \draw[arrow] (x03) -- (x0) node[midway, fill=white, text=int2_color, sloped] {$\mathbf{w_{03}}$};
    \end{tikzpicture}
}

\newcommand{\MixQExample}{
    \begin{tikzpicture}[node distance=4em and 1.25em]
      \node[circnode, fill=int4_color, text=white] (x2) {$x_2$};
      \node[circnode, fill=int8_color, below left = of x2, text=white] (x1) {$x_1$};
      \node[circnode, fill=int8_color, below right = of x2, text=white] (x0) {$x_0$};
      \node[circnode, fill=int2_color, below left = of x1, text=white] (x10) {$x_0$};
      \node[circnode, fill=int2_color, below = of x1, text=white] (x12) {$x_2$};
      \node[circnode, fill=int2_color, below left = of x0, text=white] (x01) {$x_1$};
      \node[circnode, fill=int2_color, below = of x0, text=white] (x02) {$x_2$};
      \node[circnode, fill=int2_color, below right = of x0, text=white] (x03) {$x_3$};
    
      \draw[arrow] (x1) -- (x2) node[midway, fill=white, text=int2_color, sloped] {$\mathbf{w_{12}}$};
      \draw[arrow] (x0) -- (x2) node[midway, fill=white, text=int2_color, sloped] {$\mathbf{w_{02}}$};
      \draw[arrow] (x10) -- (x1) node[midway, fill=white, text=int1_color, sloped] {$\mathbf{w_{01}}$};
      \draw[arrow] (x12) -- (x1) node[midway, fill=white, text=int1_color, sloped] {$\mathbf{w_{12}}$};
      \draw[arrow] (x01) -- (x0) node[midway, fill=white, text=int1_color, sloped] {$\mathbf{w_{01}}$};
      \draw[arrow] (x02) -- (x0) node[midway, fill=white, text=int1_color, sloped] {$\mathbf{w_{02}}$};
      \draw[arrow] (x03) -- (x0) node[midway, fill=white, text=int1_color, sloped] {$\mathbf{w_{03}}$};
    \end{tikzpicture}
}

\begin{center}
    \Legend    
\end{center}
\resizebox{\linewidth}{!}{
    \captionsetup[subfigure]{font=Large}
    \begin{tabular}{c|c|c}
        \begin{subfigure}[t]{0.6\linewidth}
            \centering
            \DQExample
            \caption{\Large DQ~\cite{tailor2021degreequant}}
        \end{subfigure} &
        \begin{subfigure}[t]{0.6\linewidth}
            \centering
            \AQExample
            \caption{\Large A$^2$Q~\cite{zhu2023rm}}
        \end{subfigure} &
        \begin{subfigure}[t]{0.6\linewidth}
            \centering
            \MixQExample
            \caption{\Large MixQ-GNN}
        \end{subfigure}
    \end{tabular}
}
        \caption{An example of unfolding tree quantization using DQ, A$^2$Q, and MixQ-GNN methods. Each color represents a different bit-width (Bit$_i$) for node or edge weights.}
        \label{figure:unfolding_tree_dq_vs_a2q_vs_mixq}
    \end{figure}

    \textbf{Degree Quantization (DQ)}~\cite{tailor2021degreequant} focuses on identifying quantized aggregation in GNN layers as a significant source of numerical error, particularly for nodes with high in-degrees. It stochastically applies full precision to high in-degree nodes during training via a Bernoulli distribution. Percentile-based quantization range determination reduces output aggregation variance. DQ demonstrates efficiency, with INT8 quantized architecture achieving prediction performance close to FP32 architecture, and INT4 architecture surpassing quantization-aware training baselines~\cite{tailor2021degreequant}. However, its reliance on node in-degree ties it to the graph’s structure. Furthermore, computing percentile statistics for quantized values per iteration makes it slower than quantization-aware training.
    
    \textbf{Aggregation Aware Quantization (A$^2$Q)}~\cite{zhu2023rm} quantizes node features using learnable parameters for quantization scale and bit-width, assigning different bit-widths to each node. To enhance compression, a memory size penalty term is introduced. A local gradient method uses quantization error to address semi-supervised training challenges. For graph-level tasks, a nearest neighbor strategy handles varying input nodes by learning a fixed number of quantization parameters and selecting the appropriate ones for unseen graphs~\cite{zhu2023rm}. Nodes are represented with varying bit-widths, leading to heterogeneous feature matrices. Although theoretical feasibility, this method is computationally inefficient. This inefficiency arises from input-dependent coupling of quantization parameters. This counteracts quantization's primary objective and increases the total number of parameters within the architecture. Six distinct learning rates for parameter groups complicate hyper-parameter optimization. Quantization of edge weights was also derived to be unnecessary in GCNs due to fusion with normalized diagonal matrices in quantizer scales~\cite{zhu2023rm}. However, this finding is limited to the GCN and is not generalized to other architectures. The derivation is constrained by assuming that the activation function must be ReLU and the edge weights must be positive, constraints that are specific to GCNs and not applicable to generic MPNNs.
    Moreover, A$^2$Q's reliance on FPGA design capabilities~\cite{Zhu2023MEGAAM} for handling dynamic bit-widths limits its applicability across current hardware~\cite{andersch2022nvidia}.
    
    \begin{table}
    \centering
    \resizebox{\linewidth}{!}{
        \begin{tabular}{lll}
        \toprule[1.5pt]
        \textbf{Method} & \textbf{Space Complexity} & \textbf{Time Complexity} \\ 
        \midrule[1pt]
        DQ~\cite{tailor2021degreequant} & ${O}\left(l + b n f l\right)$ & $ {O}_{\text{FP32}}\left(f l\right) + {O}_{\text{INT}}\left((n^2 f + n f^2) l\right)$ \\ 
        A$^2$Q~\cite{zhu2023rm} & ${O}\left(n l + \bar{b} n f l\right)$ & ${O}_{\text{FP32}}\left(n f l\right) + {O}_{\text{INT}}\left((n^2 f + n f^2) l\right)$ \\
        MixQ-GNN & ${O}\left(l + \bar{b} n f l\right)$ & ${O}_{\text{FP32}}\left(f l\right) + {O}_{\text{INT}}\left((n^2 f + n f^2) l\right)$ \\
        \bottomrule[1.5pt]
        \end{tabular}
    }
    \caption{
    Comparison of space and time complexity for GNN quantization methods. $b$ and $\bar{b}$ are bit-width and average bit-width, $n$ is the number of nodes, $f$ is the feature dimension, $l$ is the number of layers, ${O}_{\text{INT}}$ is the integer operation complexity, and ${O}_{\text{FP32}}$ is the FP32 operation complexity.}
    \label{table:complexity_analysis}    
    \end{table}
    
    Table~\ref{table:complexity_analysis} shows the complexities for DQ, A$^2$Q, and MixQ-GNN. A$^2$Q introduces an additional factor to the architectural space and time complexity, proportional to the number of nodes in the graph, compared to DQ and MixQ-GNN.

    Figure~\ref{figure:unfolding_tree_dq_vs_a2q_vs_mixq} shows an unfolding tree for a node in an arbitrary graph. DQ (Figure~\ref{figure:unfolding_tree_dq_vs_a2q_vs_mixq}(i)) assigns uniform bit-widths to all edges and node features across the tree. A$^2$Q (Figure~\ref{figure:unfolding_tree_dq_vs_a2q_vs_mixq}(ii)) uses varying bit-widths for each part of the tree based on gradient flow during learning. MixQ-GNN (Figure~\ref{figure:unfolding_tree_dq_vs_a2q_vs_mixq}(iii)) ensures that each operation within the GNN components is performed at a consistent bit-width by fixing bit-widths at each level.
        
    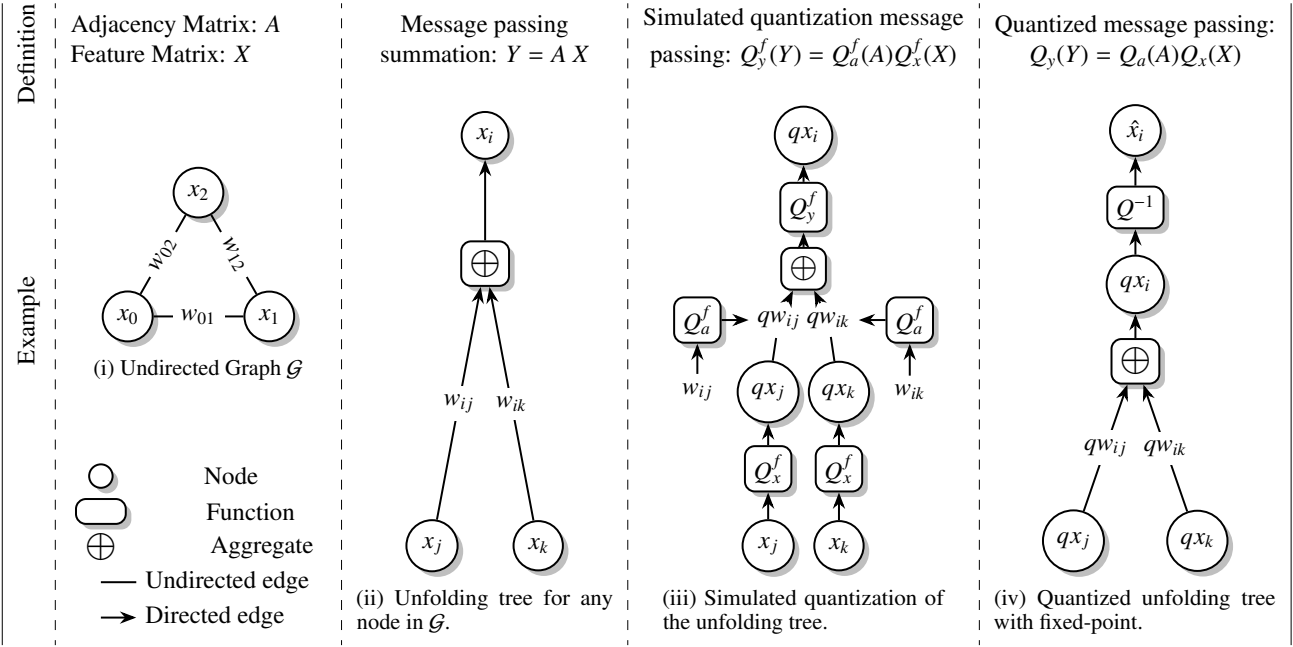
\begin{figure*}
        \centering
        \newcommand{\graphExample}{%
    \begin{tikzpicture}
      \node[circnode] (x0) at (0,0) {$x_0$};
      \node[circnode] (x1) at (2,0) {$x_1$};
      \node[circnode] (x2) at (1,1.75) {$x_2$};
    
      \draw[line] (x0) -- (x1) node[midway, fill=white] {$w_{01}$};
      \draw[line] (x0) -- (x2) node[midway, fill=white, sloped] {$w_{02}$};
      \draw[line] (x1) -- (x2) node[midway, fill=white, sloped] {$w_{12}$};
    \end{tikzpicture}
}

\newcommand{\Legend}{%
    \begin{tikzpicture}
        \node[circnode] (node1) at (0,0) {};
        \node[align=center, right=of node1] {\:\:Node};
        
        \node[draw, rectnode, minimum width=2em, minimum height=1em] (node2) at (0,-0.5) {};
        \node[align=center, right=of node2] {Function};
        
        \node (node3) at (0,-1) {$\bigoplus$};
        \node[align=center, right=of node3] {\:Aggregate};
    
        \draw[-, thick] (0,-1.5) -- +(0.5,0) node[right] {Undirected edge};
        \draw[arrow] (0,-2) -- +(0.5,0) node[right] {Directed edge};
    \end{tikzpicture}
}

\newcommand{\unfoldingTreeExample}{%
    \begin{tikzpicture}
      \def\baseX{0}
      \def\baseY{0}
      \def\verticalStep{1.45}
      \def\horizontalStep{0.5}
      \def\offset{1}
      
      \node[circnode] (xi) at (\baseX, \baseY + 4*\verticalStep) {$x_i$};
      \node[rectnode] (ai) at (\baseX, \baseY + 2.75*\verticalStep) {$\bigoplus$};
      \node[circnode] (xj) at (\baseX - 1.5*\horizontalStep, \baseY) {$x_j$};
      \node[circnode] (xk) at (\baseX + 1.5*\horizontalStep, \baseY) {$x_k$};
    
      \draw[arrow] (xj) -- (ai) node[midway, fill=white] {$w_{ij}$};
      \draw[arrow] (xk) -- (ai) node[midway, fill=white] {$w_{ik}$};
      \draw[arrow] (ai) -- (xi);
    \end{tikzpicture}
}

\newcommand{\unfoldingTreeSimulatedQuantizationExample}{%
    \begin{tikzpicture}
      \def\baseX{0}
      \def\baseY{0}
      \def\verticalStep{1.45}
      \def\horizontalStep{0.5}
      \def\offset{1}
    
      \node[circnode] (inputNodeJ) at (\baseX - \horizontalStep, \baseY) {$x_j$};
      \node[circnode] (inputNodeK) at (\baseX + \horizontalStep, \baseY) {$x_k$};
      \node[rectnode] (qInNodeJ) at (\baseX - \horizontalStep, \baseY + 0.75*\verticalStep) {$Q^f_{x}$};
      \node[rectnode] (qInNodeK) at (\baseX + \horizontalStep, \baseY + 0.75*\verticalStep) {$Q^f_{x}$};
      \node[circnode] (qTimesInputNodeJ) at (\baseX - \horizontalStep, \baseY + 1.5*\verticalStep) {$qx_j$};
      \node[circnode] (qTimesInputNodeK) at (\baseX + \horizontalStep, \baseY + 1.5*\verticalStep) {$qx_k$};
      \node[rectnode] (qOutNode) at (\baseX, \baseY + 3.3*\verticalStep) {$Q^f_{y}$};
      \node[rectnode] (addNode) at (\baseX, \baseY + 2.7*\verticalStep) {\footnotesize $\bigoplus$};
      \node[circnode] (qTimesInputNodeI) at (\baseX, \baseY + 4*\verticalStep) {$qx_i$};
      \node (edgeNodeIJ) at (\baseX - \horizontalStep - \offset, \baseY + 1.5*\verticalStep) {$w_{ij}$};
      \node (edgeNodeIK) at (\baseX + \horizontalStep + \offset, \baseY + 1.5*\verticalStep) {$w_{ik}$};
      \node (edgeIJ) at (\baseX - \horizontalStep + 0.85*\offset, \baseY + 2.2*\verticalStep) {$qw_{ik}$};
      \node (edgeIK) at (\baseX + \horizontalStep - 0.85*\offset, \baseY + 2.2*\verticalStep) {$qw_{ij}$};
      \node[rectnode] (qTimesedgeNodeJ) at (\baseX - \horizontalStep - \offset, \baseY + 2.2*\verticalStep) {$Q^f_{a}$};
      \node[rectnode] (qTimesedgeNodeK) at (\baseX + \horizontalStep + \offset, \baseY + 2.2*\verticalStep) {$Q^f_{a}$};
    
      \draw[arrow] (inputNodeJ) -- (qInNodeJ);
      \draw[arrow] (qInNodeJ) -- (qTimesInputNodeJ);
      \draw[-, thick] (qTimesInputNodeJ) -- (edgeIK);
      \draw[arrow] (edgeIK) -- (addNode);
      \draw[arrow] (inputNodeK) -- (qInNodeK);
      \draw[arrow] (qInNodeK) -- (qTimesInputNodeK);
      \draw[-, thick] (qTimesInputNodeK) -- (edgeIJ);
      \draw[arrow] (edgeIJ) -- (addNode);
      \draw[arrow] (edgeNodeIJ) -- (qTimesedgeNodeJ);
      \draw[arrow] (qTimesedgeNodeJ) -- (edgeIK);
      \draw[arrow] (edgeNodeIK) -- (qTimesedgeNodeK);
      \draw[arrow] (qTimesedgeNodeK) -- (edgeIJ);
      \draw[arrow] (qOutNode) -- (qTimesInputNodeI);
      \draw[arrow] (addNode) -- (qOutNode);
    \end{tikzpicture}
}

\newcommand{\unfoldingTreeQuantizatedExample}{%
    \begin{tikzpicture}
      \def\baseX{0}
      \def\baseY{0}
      \def\verticalStep{1.45}
      \def\horizontalStep{0.5}
      \def\offset{1}
      
      \node[circnode] (xi) at (\baseX, \baseY + 4*\verticalStep) {$\hat{x_i}$};
      \node[circnode] (qxi) at (\baseX, \baseY + 2.5*\verticalStep) {$qx_i$};
      \node[rectnode] (deq) at (\baseX, \baseY + 3.25*\verticalStep) {$Q^{-1}$};
      \node[rectnode] (ai) at (\baseX, \baseY + 1.75*\verticalStep) {$\bigoplus$};
      \node[circnode] (xj) at (\baseX - 1.75*\horizontalStep, \baseY) {$qx_j$};
      \node[circnode] (xk) at (\baseX + 1.75*\horizontalStep, \baseY) {$qx_k$};
    
      \draw[arrow] (deq) -- (xi);
      \draw[arrow] (qxi) -- (deq);
      \draw[arrow] (xj) -- (ai) node[midway, fill=white] {$qw_{ij}$};
      \draw[arrow] (xk) -- (ai) node[midway, fill=white] {$qw_{ik}$};
      \draw[arrow] (ai) -- (qxi);
    \end{tikzpicture}
}

\resizebox{0.96\linewidth}{!}{
\begin{tabular}{|l:l:c:c:c|}
    \rotatebox[origin=c]{90}{Definition} & 
    \shortstack[l]{Adjacency Matrix: $A$\\Feature Matrix: $X$} &
    \shortstack{Message passing\\summation: $Y = A \: X$} & 
    \shortstack{Simulated quantization message\\passing: $Q^f_{y}(Y) = Q^f_{a}(A) Q^f_{x}(X)$} & 
    \shortstack{Quantized message passing:\\$Q_{y}(Y) = Q_{a}(A) Q_{x}(X)$} \\
    \rotatebox[origin=c]{90}{Example\hspace{-25em}} &
    \begin{subfigure}{0.2\linewidth}
        \centering
        \graphExample
        \caption{Undirected Graph $\mathcal{G}$}
        \label{figure:three_node_graph_example}
        \vspace{3em}
        \Legend
    \end{subfigure} &
    \begin{subfigure}{0.2\linewidth}
        \centering
        \unfoldingTreeExample
        \caption{Unfolding tree for any node in $\mathcal{G}$.}
        \label{figure:three_node_unfolding_tree_example}
    \end{subfigure} &
    \begin{subfigure}{0.22\linewidth}
        \centering
        \unfoldingTreeSimulatedQuantizationExample
        \caption{Simulated quantization of the unfolding tree.}
        \label{figure:three_node_unfolding_tree_simulated_quantization}
    \end{subfigure} &
    \begin{subfigure}{0.22\linewidth}
        \centering
        \unfoldingTreeQuantizatedExample
        \caption{Quantized unfolding tree with fixed-point.}
        \label{figure:three_node_unfolding_tree_quantized_quantization}
    \end{subfigure}
\end{tabular}
}
        \caption{An overview of a quantization example of one-layer message passing: (a) Complete graph $\mathcal{G}$ with three nodes and three edges. (b) Unfolding tree with depth one. (c) Simulated quantization for the unfolding tree with three quantizers: $Q_{x}$, $Q_{a}$, and $Q_{y}$. (d) Quantized inference for the unfolding tree with dequantizer as output.}
        \label{figure:quantized_message_passing_network}
    \end{figure*}

    \textbf{Differentiable Architecture Search} introduces the concept of relaxing the discrete search space into a continuous space, known as continuous relaxation, to allow gradient flow over the discrete space. Giving $\mathcal{O}$ as the set of discrete candidate functions. The relaxation is achieved by assigning continuous probabilities to all possible connections in $\mathcal{O}$, as shown in Equation~(\ref{equation:dart_continuous_relaxation}), where $\bar{o}^{(i,j)} \in \mathcal{O}$ represents the mixed operation applied between two discrete options $i$ and $j$, and $\alpha$ is a learnable parameter for each connection~\cite{liu2018darts}.
    \begin{align}
    \bar{o}^{(i,j)}(x) = \sum_{o \in \mathcal{O}} \frac{e^{(\alpha_o^{(i,j)})}}{\sum_{o' \in \mathcal{O}} e^{(\alpha_{o'}^{(i,j)})}} o(x)
    \label{equation:dart_continuous_relaxation}
    \end{align}
    The work of~\citep{Cai2020RethinkingDS} introduces the concept of mixed precision through continuous relaxation for image classification. Additionally,~\citep{Koryakovskiy2023OneShotMF} uses continuous relaxation to propose a one-shot method for identifying subsets of the quantization configurations on the Pareto front for image classification and super-resolution tasks. However, both works are limited to basic linear and convolution-based architectures.
    
    To the best of our knowledge, no previous work addresses how to choose a different precision for each component in GNNs. Our work uniquely contributes by introducing a theorem for an efficient quantized message passing framework and developing the MixQ-GNN framework, which innovatively optimizes bit-width selections across GNN components.

\section{Methodology}    
    We provide an overview of quantization-aware training integration with the message passing schema in Figure~\ref{figure:quantized_message_passing_network}. 
    We illustrate over a complete graph example of three nodes with edge weights on every edge in Figure~\ref{figure:quantized_message_passing_network}\hyperref[figure:three_node_graph_example]{(i)}. 
    The unfolding tree for any node ($x_i$) within the graph is shown in Figure~\ref{figure:quantized_message_passing_network}\hyperref[figure:three_node_unfolding_tree_example]{(ii)}, where each node gathers messages (features and edge weights) from its neighbors, performs aggregation $\bigoplus$ over them, and updates the target node feature.
    In Figure~\ref{figure:quantized_message_passing_network}\hyperref[figure:three_node_unfolding_tree_simulated_quantization]{(iii)}, the quantization-aware training for the unfolding tree is illustrated, where the features of the 1$^\text{st}$ hop neighbors ($x_j, x_k$) are fake quantized using input quantizer $Q^f_{x}$, the edge weights ($w_{ij}, w_{ik}$) are fake quantized by $Q^f_{a}$, and the output of the aggregation function is fake quantized using $Q^f_{y}$. It is worth mentioning that any additional function introduced to the message passing schema will require a corresponding fake quantizer. 
    For the backward-pass the arrows in Figure~\ref{figure:quantized_message_passing_network}\hyperref[figure:three_node_unfolding_tree_simulated_quantization]{(iii)} would be reversed, and for non-differentiable functions STE would be used.
    If $Q^f_{x}$ determines whether the input is quantized based on the node degree, this would be equivalent to DQ~\cite{tailor2021degreequant}. If the quantizer learns different bit-widths for each node based on memory size penalties, the schema would be equivalent to A$^2$Q~\cite{zhu2023rm}. 
    In Figure~\ref{figure:quantized_message_passing_network}\hyperref[figure:three_node_unfolding_tree_quantized_quantization]{(iv)}, the inference architecture for the quantized message passing schema is shown,  where the scales and zero-points for $Q^f_{x}$, $Q^f_{a}$, and $Q^f_{y}$ can be fused into a single post-processing step. 
    The fusion can be performed by assuming that message gathering is executed through sparse matrix multiplication between the adjacency matrix $A$ and the feature matrix $X$. 
    Accordingly, Theorem~\ref{theorem:fused_quantized_message_passing} facilitates the execution of message gathering using integers instead of floats. 
    \begin{theorem}[Quantized Message Passing Schema]
    \label{theorem:fused_quantized_message_passing}
        Given the adjacency matrix $A \in \mathbb{R}^{n \times n}$ and the features matrix $X \in \mathbb{R}^{n \times f}$, with their quantized forms $Q_a(A)$ and $Q_x(X)$ respectively, and quantization parameter vectors $\{S_{a}, Z_{a}\}$, $\{S_{x}, Z_{x}\}$, $\{S_{y}, Z_{y}\}$, the quantized product $Q_y(AX)$ can be efficiently computed using the quantized forms of $A$ and $X$ to perform sparse-dense integer matrix multiplication by:
        $$ Q_y(AX) = C_1 \odot Q_a(A) Q_x(X) \odot C_2 + C_3 $$
        where $C_{1}$, $C_{2}$, and $C_{3}$ are either constants or involve operations with vectors, both of which are efficient to compute and do not create a bottleneck by definition.
        
        Such that $C_{1} = S_{a}$, and $C_{2} = S_{x} \oslash S_{y}$ are scale vectors, and $C_{3} = -\left(Q_{a}(A) \odot S_{a}\right) \mathbb{1}_n \left(Z_{x} \odot S_{x}\right)^\top \oslash S_y \mathbb{1}_n $ $ \left(Z_{a} \odot S_{a}\right)^\top \left(Q_{x}(X) \odot S_{x}\right) \oslash S_y + \left(Z_{a} \odot S_{a}\right) \left(Z_{x} \odot S_{x}\right)^\top \oslash S_y + Z_{y}$ is the zero-point offset.
    \end{theorem}    
    In cases of multi-layer message passing where there are multiple hops to aggregate information for each node, the quantization of each output hop is not recommended since the output of message passing layer will be quantized again in the next layer by the input quantizer. In this case, the quantization parameters for the output can be set to $S_{y} = \mathbb{1}$, and $Z_{y} = \mathbb{o}$, and the space and time complexity align with the complexities presented in Table~{\ref{table:complexity_analysis}}.

    Given $Q(\cdot)$ quantization function, $Q^{-1}(\cdot)$ de-quantization function, and $Q_a(A)$, $Q_x(X)$, $Q_y(Y)$ the quantized versions of matrices $A$, $X$, and $Y$, respectively, where $Y = AX$ is the aggregated messages.
    \begin{align*}
            & \because Q^f(\cdot) = Q^{-1}(Q(\cdot)) = \left(Q(\cdot) - Z\right) \odot S \\
            & \therefore Q^f_{a}(A) Q^f_{x}(X) \\
            & \qquad =  {\small \left(Q_{a}(A) \odot S_{a} - Z_{a} \odot S_{a} \right) \left(Q_{x}(X) \odot S_{x} - Z_{x} \odot S_{x}\right)} \\
            & \qquad =   \left(Q_{a}(A) \odot S_{a} \right) \left( Q_{x}(X) \odot S_{x} \right) \\
            & \qquad \quad -   \left(Q_{a}(A) \odot S_{a} \right) \mathbb{1}_n \left( Z_{x} \odot S_{x} \right)^\top \\
            & \qquad \quad -   \mathbb{1}_n \left(Z_{a} \odot S_{a} \right)^\top \left( Q_{x}(X) \odot S_{x} \right) \\
            & \qquad \quad +   \left(Z_{a} \odot S_{a} \right) \left( Z_{x} \odot S_{x}\right)^\top \\
            & \text{Let } N = \left(Q_{a}(A) \odot S_{a} Q_{x}(X) \odot S_{x} \right) \\ 
           & \text{By using Einstein's notation:} \\
            & \therefore N_{i, j} = \sum_{k} Q_{a}(A)_{i, k} \cdot S_{a_i} \cdot Q_{x}(X)_{k, j} \cdot S_{x_j} \\
            & \qquad \quad = \sum_{k} S_{a_i} \cdot Q_{a}(A)_{i, k} \cdot Q_{x}(X)_{k, j} \cdot S_{x_j} \\
            & \text{Set } \mathcal{D}_{a} \text{, and } \mathcal{D}_{x} \text{ as diagonal matrices of vectors } S_{a} \text{, and }  S_{x}\\
            & \therefore N = \mathcal{D}_{a}Q_{a}(A) Q_{x}(X) \mathcal{D}_{x}\\
            & \therefore Q^f_{a}(A) Q^f_{x}(X) =  \mathcal{D}_{a}Q_{a}(A) Q_{x}(X) \mathcal{D}_{x}\\
            & \qquad \qquad \qquad  \qquad -   \left(Q_{a}(A) \odot S_{a} \right) \mathbb{1}_n \left( Z_{x} \odot S_{x} \right)^\top \\
            & \qquad \qquad \qquad  \qquad -   \mathbb{1}_n \left(Z_{a} \odot S_{a} \right)^\top \left( Q_{x}(X) \odot S_{x} \right) \\
            & \qquad \qquad \qquad  \qquad +   \left(Z_{a} \odot S_{a} \right) \left( Z_{x} \odot S_{x}\right)^\top \\
    \end{align*}
    \begin{align*}
            & \because Q^f_{y}(Y) = \left(Q_{y}(Y) - Z_{y}\right) \odot S_{y} \\
            & \because Q^f_{y}(Y) = Q^f_{a}(A) Q^f_{x}(X) \\
            & \therefore \left(Q_{y}(Y) - Z_{y}\right) \odot S_{y} =  \mathcal{D}_{a}Q_{a}(A) Q_{x}(X) \mathcal{D}_{x}\\
            & \qquad \qquad \qquad \qquad \qquad -   \left(Q_{a}(A) \odot S_{a} \right) \mathbb{1}_n \left( Z_{x} \odot S_{x} \right)^\top\\
            & \qquad \qquad \qquad \qquad -   \mathbb{1}_n \left(Z_{a} \odot S_{a} \right)^\top \left( Q_{x}(X) \odot S_{x} \right) \\
            & \qquad \qquad \qquad \qquad \qquad +   \left(Z_{a} \odot S_{a} \right) \left( Z_{x} \odot S_{x}\right)^\top \\
            & \therefore Q_{y}(AX) =  S_{a} \odot \underline{Q_{a}(A) Q_{x}(X)} \odot S_{x} \odot S^{-1}_{y}\\
            & \qquad \qquad \qquad -   \left(Q_{a}(A) \odot S_{a} \right) \mathbb{1}_n \left( Z_{x} \odot S_{x} \right)^\top \odot S^{-1}_{y}\\
            & \qquad \qquad \qquad -   \mathbb{1}_n \left(Z_{a} \odot S_{a} \right)^\top \left( Q_{x}(X) \odot S_{x} \right) \odot S^{-1}_{y}\\
            & \qquad \qquad \qquad +   \left(\left(Z_{a} \odot S_{a} \right) \left( Z_{x} \odot S_{x}\right)^\top \odot S^{-1}_{y} \right) + Z_{y}
    \end{align*}
    $\therefore Q_{y}(AX)$ is computed in quantized format based on $Q_{a}(A) Q_{x}(X)$.

    The proof demonstrates the correctness of the fused quantized message passing $AX$, showing that the quantized output $Q_{y}(AX)$ is derived from the quantized forms of $Q_{x}(X)$ and $Q_{a}(A)$.
    We verified Theorem~\ref{theorem:fused_quantized_message_passing} in practice on GCN and GIN architectures within the code implementation using arbitrary graphs. The tests are available in the repository at \href{https://github.com/SamirMoustafa/MixQ/tree/main/test/test\_graph\_conv\_module.py}{\texttt{MixQ/test/test\_graph\_conv\_module.py}} and \href{https://github.com/SamirMoustafa/MixQ/tree/main/test/test\_graph\_iso\_module.py}{\texttt{MixQ/test/test\_graph\_iso\_module.py}}.
    
    Furthermore, the quantized sparse-dense matrix multiplication between $Q_{x}(X)$ and $Q_{a}(A)$ can be computed efficiently~\cite{Gale2020SparseGK, Li2022EfficientQS, Wang2021QGTCAQ}, reducing the primary bottleneck in GNNs.

    For example, a two-layer GCN can be defined in FP32 as $\hat{A} \: \sigma(\hat{A} X \Theta^0) \: \Theta^1$. When applying the quantized message-passing scheme from Theorem~{\ref{theorem:fused_quantized_message_passing}}, the quantized message-passing two-layer GCN is defined as $Q_{y_1}\left(\hat{A} \: \sigma\left(Q_{y_0}(\hat{A} X) \Theta^0\right)\right) \Theta^1$, where each sparse-dense matrix multiplication is performed using integer operations, corresponding to the multiplications of $\hat{A}$ with $X$, and $\hat{A}$ with the output from the first layer $\sigma\left(Q_{y_0}(\hat{A} X) \Theta^0\right)$.
    
    Given the quantized message passing schema and the proposed quantization-aware training for basic neural network functions such as linear transformation, batch normalization, etc.~\cite{Benoit2018Quantization, Nagel2021AWP}, all the components for building a quantized MPNN are defined. 

    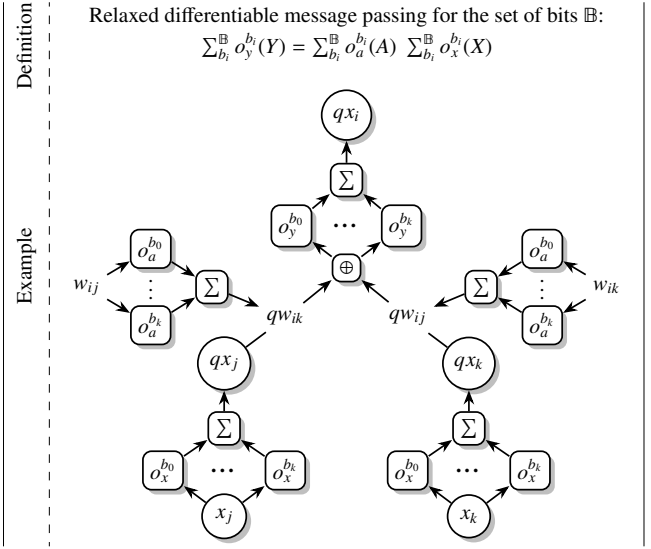
\begin{figure}
        \centering
        \resizebox{\linewidth}{!}{\newcommand{\relaxedMessagePassing}{%
    \begin{tikzpicture}
        \def\baseX{0}
        \def\baseY{0}
        \def\verticalStep{1.25}
        \def\horizontalStep{2}
        \def\offset{3}
        
        \node[circnode] (inputNodeJ) at (\baseX - \horizontalStep, \baseY - 0.5*\verticalStep) {$x_j$};
        \node[circnode] (inputNodeK) at (\baseX + \horizontalStep, \baseY - 0.5*\verticalStep) {$x_k$};
        \node[rectnode] (oInNodeJ1) at (\baseX - \horizontalStep - 0.5*\horizontalStep, \baseY + 0.1*\verticalStep) {$o^{b_0}_{x}$};
        \node (oInNodeJ1Dots) at (\baseX - \horizontalStep , \baseY + 0.1*\verticalStep) {$\textbf{\dots}$};
        \node[rectnode] (oInNodeJ2) at (\baseX - \horizontalStep + 0.5*\horizontalStep, \baseY + 0.1*\verticalStep) {$o^{b_k}_{x}$};
        \node[rectnode] (sumXjO) at (\baseX - \horizontalStep, \baseY + 0.65*\verticalStep) {$\sum$};
        \node[rectnode] (oInNodeK1) at (\baseX + \horizontalStep - 0.5*\horizontalStep, \baseY + 0.1*\verticalStep) {$o^{b_0}_{x}$};
        \node (oInNodeK2Dots) at (\baseX + \horizontalStep, \baseY + 0.1*\verticalStep) {$\textbf{\dots}$};
        \node[rectnode] (oInNodeK2) at (\baseX + \horizontalStep + 0.5*\horizontalStep, \baseY + 0.1*\verticalStep) {$o^{b_k}_{x}$};
        \node[rectnode] (sumXkO) at (\baseX + \horizontalStep, \baseY + 0.65*\verticalStep) {$\sum$};
        \node[circnode] (qTimesInputNodeJ) at (\baseX - \horizontalStep, \baseY + 1.5*\verticalStep) {$qx_j$};
        \node[circnode] (qTimesInputNodeK) at (\baseX + \horizontalStep, \baseY + 1.5*\verticalStep) {$qx_k$};
        \node[rectnode] (oOutNode1) at (\baseX + 0.3*\offset, \baseY + 3.3*\verticalStep) {$o^{b_k}_{y}$};
        \node (oOutNode1Dots) at (\baseX, \baseY + 3.3*\verticalStep) {$\textbf{\dots}$};
        \node[rectnode] (oOutNode2) at (\baseX - 0.3*\offset, \baseY + 3.3*\verticalStep) {$o^{b_0}_{y}$};
        \node[rectnode] (addNode) at (\baseX, \baseY + 2.75*\verticalStep) {$\oplus$};
        \node[circnode] (qTimesInputNodeI) at (\baseX, \baseY + 4.75*\verticalStep) {$qx_i$};
        \node[rectnode] (sumXiO) at (\baseX, \baseY + 3.9*\verticalStep) {$\sum$};
        \node (edgeNodeIJ) at (\baseX - \horizontalStep - 0.75*\offset, \baseY + 2.5*\verticalStep) {$w_{ij}$};
        \node (edgeNodeIK) at (\baseX + \horizontalStep + 0.75*\offset, \baseY + 2.5*\verticalStep) {$w_{ik}$};
        \node (edgeIJ) at (\baseX - \horizontalStep + \offset, \baseY + 2.1*\verticalStep) {$qw_{ij}$};
        \node (edgeIK) at (\baseX + \horizontalStep - \offset, \baseY + 2.1*\verticalStep) {$qw_{ik}$};
        \node[rectnode] (oTimesedgeNodeJ1) at (\baseX - \horizontalStep - 0.4*\offset, \baseY + 2*\verticalStep) {$o^{b_k}_{a}$};
        \node[rectnode] (oTimesedgeNodeJ2) at (\baseX - \horizontalStep - 0.4*\offset, \baseY + 3*\verticalStep) {$o^{b_0}_{a}$};
        \node (oTimesedgeNodeJ2Dosts) at (\baseX - \horizontalStep - 0.4*\offset, \baseY + 2.55*\verticalStep) {$\textbf{\vdots}$};
        \node[rectnode] (sumEleft) at (\baseX - 1.1*\horizontalStep, \baseY + 2.5*\verticalStep) {$\sum$};
        \node[rectnode] (oTimesedgeNodeK1) at (\baseX + \horizontalStep + 0.4*\offset, \baseY + 2*\verticalStep) {$o^{b_k}_{a}$};
        \node (oTimesedgeNodeJ2Dosts) at (\baseX + \horizontalStep + 0.4*\offset, \baseY + 2.55*\verticalStep) {$\textbf{\vdots}$};
        \node[rectnode] (oTimesedgeNodeK2) at (\baseX + \horizontalStep + 0.4*\offset, \baseY + 3*\verticalStep) {$o^{b_0}_{a}$};
        \node[rectnode] (sumEright) at (\baseX + 1.1*\horizontalStep, \baseY + 2.5*\verticalStep) {$\sum$};
    
        \draw[arrow] (inputNodeJ) -- (oInNodeJ1);
        \draw[arrow] (inputNodeJ) -- (oInNodeJ2);
        \draw[arrow] (oInNodeJ1) -- (sumXjO);
        \draw[arrow] (oInNodeJ2) -- (sumXjO);
        \draw[arrow] (sumXjO) -- (qTimesInputNodeJ);
        \draw[-, thick] (qTimesInputNodeJ) -- (edgeIK);
        \draw[arrow] (edgeIK) -- (addNode);
        \draw[arrow] (inputNodeK) -- (oInNodeK1);
        \draw[arrow] (inputNodeK) -- (oInNodeK2);
        \draw[arrow] (oInNodeK1) -- (sumXkO);
        \draw[arrow] (oInNodeK2) -- (sumXkO);
        \draw[arrow] (sumXkO) -- (qTimesInputNodeK);
        \draw[-, thick] (qTimesInputNodeK) -- (edgeIJ);
        \draw[arrow] (edgeIJ) -- (addNode);
        \draw[arrow] (edgeNodeIJ) -- (oTimesedgeNodeJ1);
        \draw[arrow] (edgeNodeIJ) -- (oTimesedgeNodeJ2);
        \draw[arrow] (oTimesedgeNodeJ1) -- (sumEleft);
        \draw[arrow] (oTimesedgeNodeJ2) -- (sumEleft);
        \draw[arrow] (sumEleft) -- (edgeIK);
        \draw[arrow] (edgeNodeIK) -- (oTimesedgeNodeK1);
        \draw[arrow] (edgeNodeIK) -- (oTimesedgeNodeK2);
        \draw[arrow] (oTimesedgeNodeK1) -- (sumEright);
        \draw[arrow] (oTimesedgeNodeK2) -- (sumEright);
        \draw[arrow] (sumEright) -- (edgeIJ);
        \draw[arrow] (sumXiO) -- (qTimesInputNodeI);
        \draw[arrow] (oOutNode1) -- (sumXiO);
        \draw[arrow] (oOutNode2) -- (sumXiO);
        \draw[arrow] (addNode) -- (oOutNode1);
        \draw[arrow] (addNode) -- (oOutNode2);
    \end{tikzpicture}
}

\resizebox{\linewidth}{!}{
\begin{tabular}{|l:c|}
    \rotatebox[origin=c]{90}{Definition} & 
    \shortstack{Relaxed differentiable message passing for the set of bits $\mathbb{B}$:\\$\sum_{b_i}^{\mathbb{B}} o^{b_i}_{y}(Y) = \sum_{b_i}^{\mathbb{B}} o^{b_i}_{a}(A) \; \sum_{b_i}^{\mathbb{B}} o^{b_i}_{x}(X)$} \\
    \rotatebox[origin=c]{90}{Example\hspace{-25em}} &
    \begin{subfigure}{1.05\linewidth}
        \centering
        \relaxedMessagePassing
    \end{subfigure}
\end{tabular}
}}
        \caption{Relaxed differentiable architecture for simulated quantization of message passing. Each relaxed connection $o^{b_i}$ contains a different quantizer which is characterized by a different bit-width $b_i \in \mathbb{B}$. Example follows Figure~{\ref{figure:quantized_message_passing_network}\hyperref[figure:three_node_graph_example]{(i)}}'s graph.}
        \label{figure:relaxed_message_passing_network}
    \end{figure}

    \subsection{\small Relaxed Message Passing for Mixed Precision Quantization}
    Given the fact that each quantizer in the message passing and the linear transformation can represent different bit-widths, we introduce MixQ-GNN, which allows each of the architecture's components to be quantized into $k$ different bit-widths. Each bit-width $b_i \in \mathbb{B}$ has its own quantizer $Q^{b_i}$, where $\mathbb{B}$ is the set of choices for the bit-widths, and $k=|\mathbb{B}|$. Since there are $k$ discrete choices for the quantizer, a continuous relaxation is applied by performing Equation~(\ref{equation:mixq_continuous_relaxation}) on each option. $\alpha$'s are the relaxed learnable parameters that are tuned by backpropagation to minimize the loss function.
    \begin{equation}
        o^{b_i}(x) = \frac{e^{\alpha_i}}{\sum_{j \in |\mathbb{B}|} e^{\alpha_j}} Q^{b_i}(x)
        \label{equation:mixq_continuous_relaxation}
    \end{equation}
    By summing over the relaxed connections $o^{b_i}$, where $i \in \{0, \dots, k\}$, we produce the fake quantization output during simulated quantization for each quantizer's $Q^f_{x}$, $Q^f_{a}$, and $Q^f_{y}$, as shown in Figure~\ref{figure:relaxed_message_passing_network}. This results in a differentiable relaxed message passing schema. This schema can be trained on graph data to adjust the relaxed probabilities derived from the softmax of the $\alpha$'s, thereby determining the effective bit-width for each function in the message passing schema. 
    
    MixQ-GNN is primarily composed of two key components: relaxed message passing and relaxed linear transformation. 
    By integrating these components, MixQ-GNN constructs a relaxed architecture, forming the core structure of the network that can be trained to find the effective bit-width for each quantizer. Training the relaxed MPNNs to minimize only the loss that is based on the node or graph level tasks would lead to a higher probability for wider bit-widths, as they can represent a wider range of numbers. Consequently, this would make high $\alpha$ values contribute more to the prediction and provide smoother representations for intermediate functions. Therefore, a penalty function is introduced in addition to the main loss function. In the optimization problem (\ref{equation:mixq_optimization_function}), $\mathcal{L}$ is the loss function that measures the architecture's training error\footnote{We use the term the training error generically, as our approach can integrate any differentiable loss function. However, in Section~\ref{section:experiments}, all experiments are conducted using the binary- or cross-entropy loss.}
    over the node or graph level tasks, $\mathcal{A}$ is the MPNN architecture, $y$ is the desired output for the task on graph $\mathcal{G}$, and $\mathcal{T}(\mathcal{A}(\mathcal{G}))$ represents the set of all tensors $T_i$ produced during the inference of $\mathcal{A}(\mathcal{G})$. $\mathcal{C}$ is the penalty function, and $\gamma$ is the threshold constraint that bounds $\mathcal{C}(\cdot)$ for all $T_i$ in the set $\mathcal{T}(\mathcal{A}(\mathcal{G}))$.
    \begin{align}
        \label{equation:mixq_optimization_function}
        \min \;          & \mathcal{L}\left(\mathcal{A}(\mathcal{G}), y\right) \\
        \text{s.t.} \;   & \Sigma_i \mathcal{C}({T}_i) \leq \gamma \;  \mid \forall {T}_i \in \mathcal{T}(\mathcal{A}(\mathcal{G})) \nonumber
    \end{align}
    By applying the Lagrange multiplier over Equation~(\ref{equation:mixq_optimization_function}), we can formulate the total loss function as $\mathcal{L}\left(\mathcal{A}(\mathcal{G}), y\right) + \lambda \Sigma_i \mathcal{C}({T}_i)$, where $\lambda$ is a hyper-parameter that is tuned to control the bit-width within the MPNN architecture.
    We define $\mathcal{C}(\cdot)$ as a function that operates over each tensor produced during the inference stage of the architecture $\mathcal{A}$, where we consider the sum of the relaxed bit-width probabilities multiplied by the number of elements as in Equation~(\ref{equation:mixq_penalty_function}), where $\vert T \vert$ is the number of elements in the tensor $T$.
    \begin{equation}
        \mathcal{C}(T) = \sum_{i=1}^{|\mathbb{B}|} \left( \frac{b_i \: e^{\alpha_i}}{\sum_{j=1}^{k} e^{\alpha_j}} \right) \vert T \vert
        \label{equation:mixq_penalty_function}
    \end{equation}
    $\mathcal{C}(\cdot)$ is normalized from bits to MB by dividing it by a factor of $1024 \times 8$. The metric is a differentiable function that reflects the memory usage of the architecture.
    The derivative of $\mathcal{C}(\cdot)$ with respect to a learnable parameters instance $\alpha_i$ is expressed as $\nabla_{\alpha_i}\mathcal{C} = \frac{\partial \mathcal{C}}{\partial \alpha_i} = |T| \frac{e^{\alpha_i}}{\sum_{j=1}^{|\mathbb{B}|} e^{\alpha_j}} \left( b_i - \frac{\sum_{i=1}^{|\mathbb{B}|} b_i e^{\alpha_i}}{\sum_{j=1}^{|\mathbb{B}|} e^{\alpha_j}} \right)$, such that $b_i$ is the corresponding bit-width choice for $\alpha_i$. For larger bit-width, the gradient grows in magnitude, amplifying its related relaxed parameter's gradient and encouraging the optimization process to favor lower the bit-widths.

    \begin{algorithm}[htbp]
    \caption{MixQ-GNN Bit-width Selection}
    \label{algorithm:mix_q}
    \KwIn{$\mathcal{G}(V, E, X, W)$ graph, $y$ desire output for the task, $\mathbb{B}$ set of choices for bit-widths, $\lambda$ hyper-parameter, $\mathcal{A}$ message passing neural network architecture, $epochs$ number of iterations, $\eta$ learning rate.}
    \KwOut{Sequence of the bit-width $\mathcal{S}$ for the architecture $\mathcal{A}$}
    
    \SetKwProg{Fn}{Function}{:}{end}
    \Fn{Build Relaxed Architecture$(\mathcal{A}, \mathbb{B})$}{
        \ForEach{$m$: module $\in \mathcal{A}$}{
            Add output quantizers with $\mathbb{B}$ choices to $m$.
            
            \uIf{m is first module in $\mathcal{A}$}{
                Add input quantizers with $\mathbb{B}$ choices to $m$.
            }
                        
            \uIf{m is a message passing layer}{
                Add {\footnotesize$\bigoplus$} quantizers with $\mathbb{B}$ choices to $m$.
                \hfill\Comment{Similar to Figure~\ref{figure:relaxed_message_passing_network}}
            }
            
            \uIf{m has learnable parameters $\Theta$}{
                Add $\Theta$ quantizers with $\mathbb{B}$ choices to $m$.
            }
        }
        \Return Relaxed $\mathcal{A}$ architecture
    }
    
    \Fn{Find Bit-widths$(\mathcal{G}, y, \mathcal{A}, \mathbb{B}, \lambda)$}{
        $\mathcal{A}^\prime \leftarrow$ Build Relaxed Architecture$(\mathcal{A}, \mathbb{B})$

        
        \For{$i=1, \hdots, epochs$}{
            \Comment{Feed forward}
            $\bar{y} \leftarrow \mathcal{A}^\prime\left(\mathcal{G}\right)$ 
            
            $\ell \leftarrow \mathcal{L}\left(\mathcal{A}^\prime (\mathcal{G}), y\right) + \lambda \Sigma_i \mathcal{C}({T}_i)$
            
            \Comment{Backpropagation}
            \For{$\theta \in \Theta$ in $\mathcal{A}^\prime$}{
                $\theta^{(i+1)} \leftarrow \theta^{(i)} - \eta \nabla_{\theta^{(i)}}$ $\mathcal{L}\left(\mathcal{A}^\prime (\mathcal{G}), y\right)$
            }
            \For{$\alpha \in o^{j}$ in $\mathcal{A}^\prime$}{
            $\alpha^{(i+1)} \leftarrow \alpha^{(i)} - \eta\nabla_{\alpha_i} \lambda \Sigma_i \mathcal{C}({T}_i)$
            }
        }
        
        $\Gamma^\ast = \argmax_{\alpha \in m} \text{ for all } m \in \mathcal{A}^\prime$
        
        $\mathcal{S} = ( b_i \in \mathbb{B} \mid b_i \text{ corresponds to } \alpha_i \in \Gamma^\ast)$
        
        \Return $\mathcal{S}$
    }
    \end{algorithm}

    \subsection{MixQ-GNN Bit-Width Selection}
    To find the effective sequence of bit-widths, we employ a continuous relaxation approach, as detailed in Algorithm~\ref{algorithm:mix_q}, its space and time complexity is as same as training quantization-aware architecture multiplied by the number of bit-width choices ${O}(\text{full precision architecture}) \times |\mathbb{B}|$.
    
    The algorithm builds a relaxed architecture where each quantizer in it can operate at multiple bit-widths, controlled by learnable parameters $\alpha$. The process begins by constructing a relaxed version of the $\mathcal{A}$ architecture. For each module, a list of input and output quantizers with multiple bit-width choices are added. Message passing layers and those with learnable parameters receive additional quantizers. These quantizers are defined as Equation~(\ref{equation:mixq_continuous_relaxation}). The output of all quantizers within a component is then summed. This setup ensures that every part of the network (components) can flexibly adjust its precision to different bit-widths.
    Next, the continuous relaxation parameters $\alpha$ for all quantizers are initialized. These parameters are tuned during training via gradient-based optimization to find the optimal bit-width. The penalty function $\mathcal{C}(\cdot)$ is weighted by a hyper-parameter $\lambda$, which balances  the trade-off between prediction performance and efficiency, and equivalent to the threshold constraint that bounds $\mathcal{C}(\cdot)$ in Equation~(\ref{equation:mixq_optimization_function}). Backpropagation is then used to update both the network parameters and the relaxation parameters.
    For illustrative purposes, the gradients of $\Theta$ and $\alpha$'s are decoupled into two separate steps in lines \textbf{19} and \textbf{22}, respectively. This clarifies which loss component corresponds to each parameter. At the implementation level, all statistics concerning tensor sizes and their bit-width are collected dynamically during the feed forward process, and the gradient of $\ell$ is computed numerically using an automatic gradient computation package~{\cite{AnselPyTorch2Faster2024}}, and update is done in a single loop for all learnable parameters.
    
    After training, the optimal bit-widths are determined by selecting the bit-width corresponding to the highest $\alpha$ value for each quantizer. This results in a sequence of bit-widths for the architecture, ensuring that each part of the network operates at the effective precision level.

    \begin{table*}
        \centering
        \resizebox{\linewidth}{!}{
        \begin{tabular}{| p{5em} C | >{\centering\arraybackslash}p{9em} | >{\centering\arraybackslash}p{13em} | >{\centering\arraybackslash}p{10em} | >{\centering\arraybackslash}p{11em} |}
            
            \multicolumn{2}{|c|}{\textbf{(1) Example Setup}} &
            \textbf{(2) FP32 Inference} &
            \textbf{(3) Relaxation} &
            \textbf{(4) Bit-width Selection} &
            \textbf{(5) Quantized Inference} \\
            &&&&&\\
            Adjacency matrix: $\hat{A}$ &
            \multirow{2}{*}{\graphRandomExample} &
            \multirow{2}{*}{\shortstack{Aggregate \\features: $\hat{A} X$}} &
            $\hat{A}_r = \sum_{b_i}^{\mathbb{B}} o^{b_i}_{a}(\hat{A})$ &
            \multirow{2}{*}{\shortstack{$\Gamma^\ast = \argmax_{\alpha}$\\ s.t. $\Gamma = \forall \alpha \in \mathcal{M}$}} &
            \multirow{2}{*}{\shortstack{Aggregate features:\\ $\left(\hat{A}_{b_i} X_{b_i}\right)_{b_j}$}} \\
            
            \multirow{2}{*}{\shortstack{\quad\\\quad\\Feature\\matrix: $X$}} & 
            \multirow{2}{*}{\rowMajorMatrix} & 
            & 
            \shortstack{$X_r = \sum_{b_i}^{\mathbb{B}} o^{b_i}_{x}(X)$\\\:} & 
            & 
            \\
            
            & 
            & 
            \multirow{2}{*}{\shortstack{Predictions:\\$\bar{y} = (\hat{A} X)\Theta$}} & 
            \shortstack{$\Theta_r = \sum_{b_i}^{\mathbb{B}} o^{b_i}_{\theta}(\Theta)$\\\:} & 
            \multirow{2}{*}{\shortstack{$\Gamma^\ast = \{\alpha^\ast_a, \alpha^\ast_x, \alpha^\ast_\theta,$\\$\hspace{1cm}\alpha^\ast_{ax}, \alpha^\ast_{ax\theta}\}$}} & 
            \multirow{2}{*}{\shortstack{Predictions: $\bar{y}_{b_j} = $\\$\left(\hat{A}_{b_i} X_{b_i}\right)_{b_j} \Theta_{b_j}$}} \\ 
            
            &
            &
            & 
            \shortstack{$(\hat{A}_r X_r)_r = \sum_{b_i}^{\mathbb{B}} o^{b_i}_{ax}(\hat{A}_r X_r)$\\\:} & 
            & 
            \\
            
            Learnable matrix: $\Theta$ & 
            \multirow{3}{*}{\thetaZeroLearnable} & 
            \multirow{3}{*}{\shortstack{Evaluation:\\$\mathcal{L}(\bar{y}, y)$}} & 
            \shortstack{$\bar{y}_r = \sum_{b_i}^{\mathbb{B}} o^{b_i}_{ax\theta}((\hat{A}_r X_r)_r\Theta)$} & 
            \multirow{2}{*}{\shortstack{$\mathcal{S} = \{b_a, b_x, b_\theta,$\\$\hspace{1cm}b_{ax}, b_{ax\theta}\}$}} & 
            \multirow{3}{*}{\shortstack{\quad\\Evaluation:\\$\mathcal{L}(\bar{y}_{b_j}, y)$}} \\ 
            
            & 
            & 
            & 
            \multirow{2}{*}{Evaluation: $\mathcal{L}(\bar{y}, y) + \lambda\mathcal{C}(\bar{y}_r)$}
            & 
            & 
            \\ 
            
            &
            & 
            &
            &
            & 
            \\ 
        \end{tabular}
        }
        
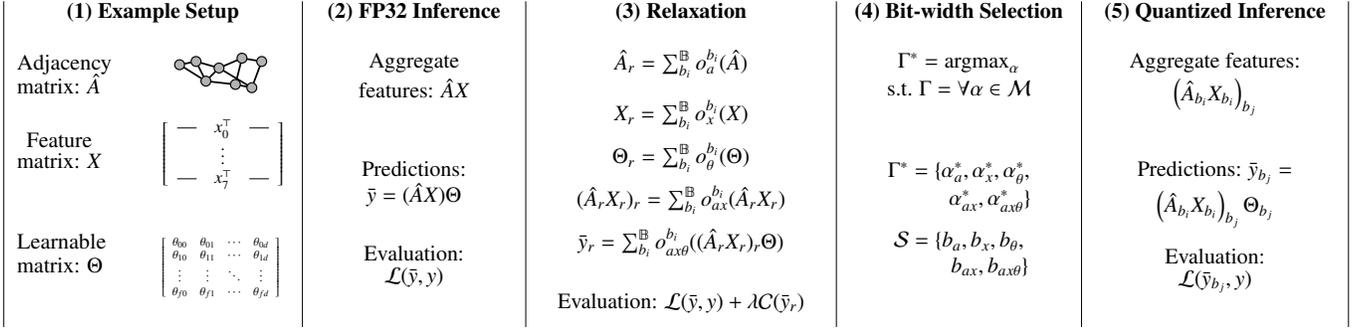
\captionof{figure}[]{MixQ-GNN with a single-layer GCN example.
            \textbf{(1) Example Setup:} Normalized adjacency matrix $\hat{A}$, node feature matrix $X$, and learnable parameter $\Theta$.
            \textbf{(2) FP32 Inference:} Messages are aggregated from neighbors via $\hat{A}$, combined with $X$, multiplied by $\Theta$ to compute predictions $\bar{y}$, and evaluated with loss $\mathcal{L}(\bar{y}, y)$.
            \textbf{(3) Relaxation:} Each matrix or operation ($\hat{A}$, $X$, $\Theta$, $\hat{A}X$, and $\bar{y}$) is relaxed using a relaxation process from Equation~{\ref{equation:mixq_continuous_relaxation}}, with the relaxed components denoted by the subscript `r'. These relaxed components form the set $\mathcal{M} = \{\hat{A}_r, X_r, \Theta_r, (\hat{A}_r X_r)_r, \bar{y}_r\}$. Evaluation sums both the original loss and the penalty term $\lambda \mathcal{C}(\bar{y}_r)$.
            \textbf{(4) Bit-width Selection:} After training, effective bit-widths are selected by selecting the highest relaxation parameters $\alpha$ for each component to form the set $\Gamma$. These bit-widths are assigned to its corresponding components in the sequence $\mathcal{S}$, determining the bit-width for each matrix and operation.
            \textbf{(5) Quantized Inference:} The quantized aggregated features $\hat{A}_{b_i} X_{b_i}$ are computed using the \textit{Quantized Message Passing Schema} from Theorem~{\ref{theorem:fused_quantized_message_passing}}. Predictions $\bar{y}_{b_j}$ are computed via quantized matrix operations and evaluated using the loss function $\mathcal{L}(\bar{y}_{b_j}, y)$.
        }
        \label{figure:mixq_gnn_overview}
    \end{table*}

    Figure~{\ref{figure:mixq_gnn_overview}} provides an illustrative example overview of a single-layer GCN with five stages, summarizing the pipeline of MixQ-GNN. Our approach begins with the relaxation stage, where the $\mathcal{C}$ term is introduced as a regularization term. In this stage, its gradient minimizes the relaxation parameters, and the bit-widths are subsequently selected based on the highest values of these relaxation parameters. Theorem~{\ref{theorem:fused_quantized_message_passing}} is then applied during the quantized inference stage.

\section{Experimental Evaluation}
\label{section:experiments}
    In this section, we evaluate our MixQ-GNN framework\footnote{Code available at \href{https://github.com/SamirMoustafa/MixQ}{\texttt{ github.com/SamirMoustafa/MixQ}}} on node classification and graph classification tasks. 
    Characteristics about the datasets used can be found in Table~\ref{table:datasets_summary}. {Cora}, {CiteSeer}, {PubMed}, {OGB-Arxiv}, and Illinois Graph Benchmark ({IGB}) are utilized for node and document classification within citation and biomedical literature networks. Social network analyses are performed using {OGB-Products}, {IMDB-B}, {IMDB-M}, {REDDIT-B}, {REDDIT-M}, and {Reddit}, focusing on movie genre classification, user interactions, and community detection. Biological graph classification is conducted with the {OGB-Protein}, {PROTEINS} and {D\&D} datasets, while the synthetic {CSL} dataset supports graph theory research through multi-class classification of graph structures. For datasets lacking node features, one-hot encoding based on node degree was applied, and Laplacian positional encoding with 50 eigenvectors was used for CSL dataset.
    Although the existing baselines~\cite{tailor2021degreequant, zhu2023rm} use either graph statistics or over parameterized methods, we demonstrate the advantages of mixed precision by using native quantization-aware training quantizers. Additionally, MixQ-GNN is flexible, allowing any quantization methodology to be integrated by simply replacing the definition of the quantization and de-quantization functions $Q$, and $Q^{-1}$ respectively.

    \subsection{Efficiency Metric}
    To evaluate the efficiency of mixed precision architectures, we use the metric Bit Operations (BitOPs) as in other works~\cite{SdqHuang22h, BayesianBitsBaalen, SearchLow8bitYang}. This metric ensures consistent computational cost comparison, instead of only relying on average bit width (Bits).
    To compute the BitOPs in the architecture $\mathcal{A}$, it is regarded as a collection of distinct modules, each comprising various functions. Each function is associated with a numerical value representing the quantity of operations it executes. 
    Each module tracks operations from initialization to completion. These operations process data of varying sizes but maintain consistent bit-widths, although distinct functions may operate with differing bit-widths. 
    The total BitOPs for a module are determined by calculating the weighted average of its bit-widths with respect to the number of operations performed per function. By summing these totals across all modules, one can derive a comprehensive measure of the architecture's BitOPs.

\begin{figure}
        \centering
        \includegraphics[width=\linewidth]{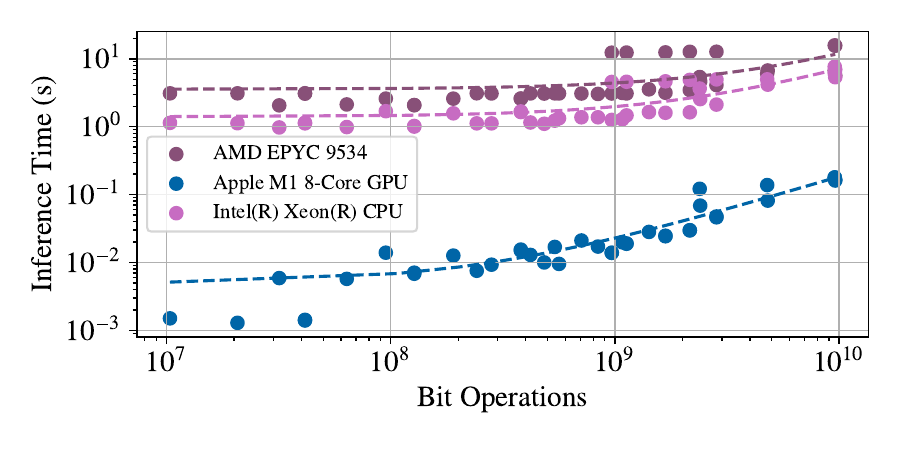}
        \caption{
        Scatter plot between BitOPs and inference times for single message passing layer, represented across three hardware platforms: AMD, Apple M1, and Intel. Each point on the scatter plot corresponds to a specific dataset's BitOPs and the corresponding inference time. Linear regression lines (dashed) indicate the general trends. Due to the logarithmic scaling of both axes, the regression lines appear curved.}
        \label{figure:message_passing_time_vs_bit_operations}
    \end{figure}

    Figure~\ref{figure:message_passing_time_vs_bit_operations} presents a scatter plot with logarithmic scales on both axes, illustrating the relationship between the number of BitOPs and inference times across three distinct hardware platforms. The plot was conducted using a different datasets and bit-widths (INT8, INT16, INT32, and FP32) for a single layer of message passing. It reveals different degrees of positive correlation. The AMD EPYC 9534 64-Core processor~\cite{amd_epyc_9534} exhibits a moderate Pearson correlation coefficient of $0.59$, indicating a noticeable relationship between the variables. The Apple M1 8-Core GPU (ARMv8.5-A)~\cite{ARM_Architecture_Manual} demonstrates a very strong correlation, with a coefficient of $0.95$. This strong correlation is attributed to the Apple M1's integrated GPU, which is effectively utilized during the message passing process, resulting in significant speedup. Additionally, the Intel Xeon processor on Google Colaboratory~\cite{Bisong2019} shows a strong correlation, possessing a coefficient of $0.70$. Based on these correlations, inference times increase proportionally as BitOPs increase across different hardware platforms. 
    The script utilized to obtain results in Figure~\ref{figure:message_passing_time_vs_bit_operations} is accessible at \href{https://github.com/SamirMoustafa/MixQ/tree/main/hardware\_speedup/message\_passing\_with\_diff\_precision.py}{\texttt{MixQ/hardware\_speedup/message\_passing \_with\_diff\_precision.py}}. It can be used to validate the analysis for different hardware configurations.

    \begin{table}
    \centering
    \resizebox{\linewidth}{!}{
    \begin{tabular}{lccccc}
        \toprule[2pt]
        \textbf{Dataset} & \boldmath\textbf{$\vert$G$\vert$} & \boldmath\textbf{$\overline{\vert V\vert}$} & \boldmath\textbf{$\overline{\vert E \vert}$} & \boldmath\textbf{$\vert X \vert$}  & \boldmath\textbf{$\vert Y \vert$} \\
        \midrule[1pt]
        \textbf{CiteSeer}~\cite{Zhilin2016RevisitingGraphEmbeddings} & 1 & 3,327 & 9,104 & 3,703 & 6\\
        \textbf{Cora}~\cite{Zhilin2016RevisitingGraphEmbeddings} & 1 & 2,708 & 10,556 & 1,433 & 7\\
        \textbf{PubMed}~\cite{Zhilin2016RevisitingGraphEmbeddings} & 1 & 19,717 & 88,648 & 500 & 3\\
        \textbf{OGB-Arxiv}~\cite{hu2020ogb} & 1 & 169,343 & 1,166,243 & 128 & 40\\
        \textbf{IGB}~{\cite{igbdatasets}} & 1 & 1,000,000 & 12,070,502	 & 1024 & 19\\
        \textbf{OGB-Proteins}~\cite{hu2020ogb} & 1 & 132,534 & 39,561,252 & 112 & 112\\
        \textbf{OGB-Products}~\cite{hu2020ogb} & 1 & 2,449,029 & 61,859,140	 & 100 & 47\\
        \textbf{Reddit}~\cite{Hamilton2017InductiveRL} & 1 & 232,965 & 114,615,892 & 602 & 41\\
        \textbf{CSL}~\cite{murphy2019relational} & 150 & 41.0 & 164.0 & - & 10\\
        \textbf{IMDB-B}~\cite{Morris2020TuDataset} & 1,000 & 19.8 & 193.1 & - & 2\\
        \textbf{PROTEINS}~\cite{Morris2020TuDataset} & 1,113 & 39.1 & 145.6 & 3 & 2\\
        \textbf{D\&D}~\cite{Morris2020TuDataset} & 1,178 & 284.3 & 715.6 & 89 & 2\\
        \textbf{REDDIT-B}~\cite{Morris2020TuDataset} & 2,000 & 429.6 & 497.7 & - & 2\\
        \textbf{REDDIT-M}~\cite{Morris2020TuDataset} & 4,999 & 508.8 & 594.9 & - & 5\\
        \bottomrule[2pt]\vspace{0.0em}
    \end{tabular}
    }
    \caption{Datasets characteristics with dataset name, number of graphs ($|G|$), average number of nodes ($\overline{|V|}$), average number of edges ($\overline{|E|}$), number of classes ($|Y|$), and feature dimension ($|X|$). ``-'' indicate not available node features.}
    \label{table:datasets_summary}
    \end{table}
    
    \subsection{Hardware and System Configuration}
        \label{section:hradware_configuration}
         Experiments were conducted on a machine with an AMD EPYC 9534 64-Core Processor ($256$ CPUs, max frequency $3.72$ GHz) running Ubuntu $22.04.4$ LTS and Python $3.11.5$. An NVIDIA A100 GPU with CUDA $12.4$.

    \subsection{Node level tasks}
    
    \begin{table}
       \begin{center}
       \resizebox{\linewidth}{!}{
       \begin{tabular}{llccc}
       
       \toprule[2pt]
       \textbf{Dataset}                       & \textbf{Method}                             & \textbf{Accuracy $\uparrow$}  & \textbf{Bits} &  \textbf{GBitOPs $\downarrow$} \\  
       \midrule[1pt]
       \multirow{7}{*}{\textbf{Cora}}         & FP32                                                & 81.5$\pm$0.7\%                        & 32     & 16.11                    \\
                                              & DQ ~\cite{tailor2021degreequant}                    & \textcolor{first}{81.7$\pm$0.7\%}     & 8      & 4.03                     \\
                                              & DQ ~\cite{tailor2021degreequant}                    & 78.3$\pm$1.7\%                        & 4      & \textcolor{second}{2.01} \\
                                              & A$^2$Q~\cite{zhu2023rm}                             & \textcolor{third}{80.9$\pm$0.6\%}     & 1.70   & 8.94                     \\ 
        \cmidrule{2-5}
                                              & MixQ $_{(\lambda=-\varepsilon)}$                    & \textcolor{second}{81.6$\pm$0.7\%}    & 7.69   & 3.95                     \\ 
                                              & MixQ $_{(\lambda=0.1)}$                             & 77.7$\pm$2.8\%                        & 5.82   & \textcolor{third}{3.35}  \\ 
                                              & MixQ $_{(\lambda=1)}$                               & 68.7$\pm$2.7\%                        & 3.84   & \textcolor{first}{1.68}  \\ 
       \midrule[1pt]
       \multirow{7}{*}{\textbf{CiteSeer}}     & FP32                                                & 71.1$\pm$0.7\%                        & 32     & 50.68                    \\
                                              & DQ ~\cite{tailor2021degreequant}                    & \textcolor{first}{71.0$\pm$0.9\%}     & 8      & 12.67                    \\
                                              & DQ ~\cite{tailor2021degreequant}                    & 66.9$\pm$2.4\%                        & 4      & \textcolor{third}{6.33}  \\
                                              & A$^2$Q~\cite{zhu2023rm}                             & \textcolor{second}{70.6$\pm$1.1\%}    & 1.87   & 8.96                     \\ 
        \cmidrule{2-5}
                                              & MixQ $_{(\lambda=-\varepsilon)}$                    & \textcolor{third}{69.0$\pm$1.1\%}     & 6.84   & 12.44                    \\ 
                                              & MixQ $_{(\lambda=0.1)}$                             & 66.5$\pm$1.8\%                        & 4.49   & \textcolor{second}{5.18} \\ 
                                              & MixQ $_{(\lambda=1)}$                               & 60.9$\pm$8.7\%                        & 3.44   & \textcolor{first}{4.23}  \\ 
       \midrule[1pt]
       \multirow{7}{*}{\textbf{PubMed}}       & FP32                                                & 78.9$\pm$0.7\%                        & 32     & 41.7                     \\
                                              & DQ                                                  & NA                                    & NA     & NA                       \\
                                              & DQ ~\cite{zhu2023rm}                                & 62.5$\pm$2.4\%                        & 4      & \textcolor{second}{5.21} \\
                                              & A$^2$Q~\cite{zhu2023rm}                             & \textcolor{second}{77.5$\pm$0.1\%}    & 1.90   & 8.94                     \\
        \cmidrule{2-5}
                                              & MixQ $_{(\lambda=-\varepsilon)}$                    & \textcolor{first}{78.3$\pm$0.2\%}     & 7.36   & 10.34                    \\ 
                                              & MixQ $_{(\lambda=0.1)}$                             & \textcolor{third}{77.3$\pm$0.7\%}     & 5.49   & \textcolor{third}{6.89}  \\ 
                                              & MixQ $_{(\lambda=1)}$                               & 71.0$\pm$1.8\%                        & 4.09   & \textcolor{first}{4.85}  \\ 
       \midrule[1pt]
       \multirow{7}{*}{\textbf{OGB-Arxiv}}    & FP32                                                & 71.7$\pm$0.3\%                        & 32     & 692.87                   \\          
                                              & DQ                                                  & NA                                    & NA     & NA                       \\
                                              & DQ~\cite{tailor2021degreequant}                                 & 65.4$\pm$3.9\%                        & 4      & \textcolor{first}{86.96} \\
                                              & A$^2$Q~\cite{zhu2023rm}\footref{note:a2q_drawback}  & \textcolor{first}{71.1$\pm$0.3\%}     & 2.65   & \textcolor{second}{141.93}\\
        \cmidrule{2-5}
                                              & MixQ $_{(\lambda=-\varepsilon)}$                    & \textcolor{second}{70.6$\pm$0.0\%}    & 8.00   & \textcolor{third}{167.50}\\ 
                                              & MixQ $_{(\lambda=0.1)}$                             & \textcolor{third}{70.0$\pm$0.0\%}     & 7.08   & 167.50                   \\ 
                                              & MixQ $_{(\lambda=1)}$                               & 69.3$\pm$0.0\%                        & 7.08   & 167.50                   \\ 
       \bottomrule[2pt]
       \end{tabular}
       }
       \end{center}
        \caption{Node classification accuracy using the GCN architecture, with the corresponding average bit-width (``Bits'') and Giga bit operations (``GBitOPs''). The \textcolor{first}{\textbf{first-}}, \textcolor{second}{\textbf{second-}} and \textcolor{third}{\textbf{third-}}best results are colored. Results from~\cite{zhu2023rm} are compared against MixQ-GNN over the mean of ten runs. The term ``NA'' indicates that the result is not available or reproducible. The bit-width search space for MixQ-GNN is $\{2, 4, 8\}$ for the datasets Cora, CiteSeer, and PubMed, and $\{4, 8\}$ for OGB-Arxiv. The value of epsilon is set to $\varepsilon = 10^{-8}$.
        }
       \label{figure:node-level-dq-vs-a2q-vs-mixq}
    \end{table}
    
    \begin{table}
       \begin{center}
       \resizebox{\linewidth}{!}{
       \begin{tabular}{lccc}
       
       \toprule[2pt]
       \textbf{Method}                       & \textbf{Accuracy $\uparrow$}  & \textbf{Bits} &  \textbf{GBitOPs $\downarrow$} \\  
       \midrule[1pt]
       MixQ$_{(\lambda=-\varepsilon)}$      & 81.6$\pm$0.7\%     & 7.69   & 3.95   \\ 
       MixQ$_{(\lambda=-\varepsilon)}$ + DQ  & \textbf{81.8$\pm$0.3\%}     & 7.69   & 4.01   \\ 
       \midrule
       MixQ$_{(\lambda=0.1)}$               & 77.7$\pm$2.8\%     & 5.82   & 3.35   \\ 
       MixQ$_{(\lambda=0.1)}$ + DQ           & \textbf{79.9$\pm$0.6\%}     & 6.02   & 3.35   \\ 
       \midrule
       MixQ$_{(\lambda=1)}$                 & 68.7$\pm$2.7\%     & 3.84   & 1.68   \\ 
       MixQ$_{(\lambda=1)}\;\;$ + DQ         & \textbf{72.3$\pm$1.2\%}     & 3.69   & 1.68   \\        
       \bottomrule[2pt]
       \end{tabular}
       }
       \end{center}
        \caption{Comparison between native quantizer and DQ~\cite{tailor2021degreequant} quantizers with MixQ-GNN on node classification with a two GCN layer architecture on the Cora dataset with ten runs. The bit-width search space is $\{2, 4, 8\}$, and $\varepsilon = 10^{-8}$.}
       \label{figure:node-level-cora-mixq-plus-dq}
    \end{table}
    
    We explore the prediction performance of the MixQ-GNN method on node level tasks using the GCN architecture for classification using the datasets Cora, CiteSeer, PubMed~\cite{Zhilin2016RevisitingGraphEmbeddings}, and OGB-Arxiv~\cite{hu2020ogb}.
    MixQ-GNN demonstrates competitive accuracy on node classification compared to other quantization techniques with the GCN architecture. As shown in Table~\ref{figure:node-level-dq-vs-a2q-vs-mixq}, MixQ-GNN selects between bit-widths $\{2, 4, 8\}$ for Cora, CiteSeer, PubMed datasets, and $\{4, 8\}$ for OGB-Arxiv, since they are the most realistic hardware choices. This allows for fair comparison with~\cite{tailor2021degreequant}.
    
    MixQ-GNN maintains proper accuracy over full precision architecture while flexibly shifting between bit-widths, balancing computational efficiency and accuracy performance. For the Cora dataset, MixQ-GNN with $\lambda = -10^{-8}$ achieves slightly higher accuracy than FP32, while on CiteSeer, PubMed, and OGB-Arxiv, it is only slightly worse. When $\lambda$ is set to $0.1$ and $1$, MixQ-GNN minimizes both average bit-width and bit operations, reducing computational cost under the given constraints. 
    In OGB-Arxiv dataset, A$^2$Q seems to have an advantage over MixQ-GNN due to over-parameterization, by using different bit-widths for each node and depending on the graph structure. Each of the $169,343$ nodes has its own scale and bit-width parameters in FP32 representation. In total, there are $2 \times 169,343$ quantization parameters, exceeding the original $164,608$ parameters found in a three-layers GCN with an FP32 architecture\footnote[{$\mathsection$}]{\label{note:a2q_drawback}Based on the A$^2$Q code, the FP32 GCN architecture requires $110,120$ learnable parameters, while their quantized GCN architecture requires $788,596$ learnable parameters. Contrary to this, three GCN layers with MixQ-GNN over OGB-Arxiv dataset require only $113,318$ learnable parameters.}, which aligns with the complexity analysis in Table~\ref{table:complexity_analysis}, where the space and time complexity of A$^2$Q has an overhead factor that depends on the graph size.

    To assess how relying on graph structure enhances MixQ-GNN's ability to improve quantized architecture accuracy, we trained a relaxed GCN layer architecture on the Cora to determine the appropriate bit-width choices. Then we use the DQ quantizer~\cite{tailor2021degreequant} to train this architecture.
    In Table~\ref{figure:node-level-cora-mixq-plus-dq}, the combination of MixQ + DQ outperformed both MixQ and DQ individually due to the effective combination chosen by MixQ-GNN and the degree-aware quantization from DQ.

    On average, MixQ-GNN achieves a reduction in bit operations of $5.5\times$ compared to the FP32 architecture across Cora, CiteSeer, PubMed, and OGB-Arxiv datasets.

    \subsubsection{MixQ-GNN versus A$^2$Q}
    
    \begin{table}
        \centering
        \resizebox{0.95\linewidth}{!}{
        \begin{tabular}{llcc}
        \toprule[2pt]
        \textbf{Dataset} & \textbf{Method} & \textbf{Accuracy $\uparrow$} & \textbf{GBitOPs $\downarrow$} \\
        \midrule[1pt]
        \multirow{2}{*}{\textbf{Cora}}      & A$^2$Q~\cite{zhu2023rm}         & 80.9$\pm$0.6\% & 8.94 \\
                                   & MixQ + DQ   & \textbf{81.8$\pm$0.3\%} & \textbf{4.01} \\
        \midrule[1pt]
        \multirow{2}{*}{\textbf{CiteSeer}}  & A$^2$Q~\cite{zhu2023rm}         & \textbf{70.6$\pm$1.1\%} & 8.96 \\
                                   & MixQ + DQ   & 66.2$\pm$1.2\%     & \textbf{6.01}   \\ 

        \midrule[1pt]
        \multirow{2}{*}{\textbf{PubMed}}    & A$^2$Q~\cite{zhu2023rm}         & 77.5$\pm$0.1\% & 8.94 \\
                                   & MixQ + DQ   & \textbf{77.6$\pm$0.3\%} & \textbf{6.88} \\
        \bottomrule[2pt]\vspace{0.0em}
        \end{tabular}
        }
        \caption{Comparison between MixQ-GNN + DQ and A$^2$Q across different datasets. Both of methods reply on the graph structure to preform the quantization for the aggregated values.}
        \label{table:mixq_vs_a2q_nodel_level_task}
    \end{table}
    
    To ensure a fair and consistent evaluation, both A$^2$Q and MixQ-GNN utilize quantization methods that leverage the underlying graph structure for aggregating values. Table~\ref{table:mixq_vs_a2q_nodel_level_task} presents a comparison of A$^2$Q against MixQ-GNN with DQ quantizer approach across three datasets, focusing on accuracy and GBitOPs. On the Cora and PubMed datasets, MixQ-GNN combined with the DQ quantizer demonstrates either an improvement or maintenance of accuracy compared to A$^2$Q, while simultaneously achieving a reduction in computational overhead by approximately half. However, on the CiteSeer dataset, MixQ-GNN experiences a slight decrease in accuracy relative to A$^2$Q. 
    
    \subsubsection{GraphSAGE Case Study}
    
    \begin{table}
        \resizebox{\linewidth}{!}{
        \centering
        \begin{tabular}{llcccc}
            \toprule[2pt]
            \textbf{Dataset}   & \textbf{Method}    & \textbf{Accuracy $\uparrow$} & \textbf{Bits} & \textbf{GigaBitOPs $\downarrow$} \\
            \midrule[1pt]
            \multirow{3}{*}{\textbf{Cora}} 
                      & FP32         & $76.7\pm0.31\%$          & 32   & 7.8        \\
                      & MixQ $_{(\lambda=0.1)}$          & $78.1\pm0.28\%$          & 6.9  & 1.94       \\
                      & MixQ $_{(\lambda=1)}$            & $75.4\pm0.73 \%$          & 4.9  & 0.9        \\
            \midrule[1pt]
            \multirow{3}{*}{\textbf{CiteSeer}} 
                      & FP32         & $65.6\pm0.72 \%$          & 32   & 19.5       \\
                      & MixQ $_{(\lambda=0.1)}$          & $65.8\pm0.61 \%$          & 6.3  & 4.2        \\
                      & MixQ $_{(\lambda=1)}$            & $66.6\pm0.86 \%$          & 4.7  & 2.1        \\
            \midrule[1pt]
            \multirow{3}{*}{\textbf{PubMed}} 
                      & FP32         & $77.9\pm0.15 \%$          & 32   & 5.6        \\
                      & MixQ $_{(\lambda=0.1)}$          & $77.8\pm0.24 \%$         & 6.9  & 1.2        \\
                      & MixQ $_{(\lambda=1)}$            & $77.9\pm0.12 \%$         & 5.4  & 0.7        \\
            \bottomrule[2pt]\vspace{0.0em}
        \end{tabular}
        }
        \caption{Node classification accuracy using the GraphSAGE architecture, with the corresponding average bit-width (``Bits'') and Giga bit operations (``GBitOPs''). The bit-width search space for MixQ-GNN is $\{2, 4, 8\}$.}
        \label{table:graph_sage_node_level_task}
    \end{table}    
    Since the primary source of quantization error in GNNs arises from the highly divergent magnitudes of aggregated values, particularly in nodes with high in-degree~\cite{tailor2021degreequant, zhu2023rm, Wang2023LowbitQF}, we can use method node sampling~\cite{Hamilton2017InductiveRL} from GraphSAGE to reduce the nodes' in-degree and minimize these errors.
    Building upon this approach, we evaluate the performance of MixQ-GNN as a standalone model without incorporating advanced quantizers. 
    Table~\ref{table:graph_sage_node_level_task} presents the node classification accuracy for the GraphSAGE architecture across three datasets.
    Across the Cora, CiteSeer, and PubMed datasets, MixQ-GNN demonstrates the ability to significantly reduce the bit-width and computational operations while maintaining or even slightly improving classification accuracy compared to FP32 baseline. This indicates that MixQ-GNN can effectively compress the model, leading to lower memory usage and faster computations without compromising performance.
    The ability to aggressively quantize with higher lambda values while sustaining minimal accuracy fluctuations highlights MixQ-GNN's robustness and adaptability across different graph structures and dataset characteristics.
    \subsubsection{Large-Scale Experiments}
    To evaluate the scalability of MixQ-GNN, we integrated MixQ-GNN with the GraphSAGE architecture and applied it to four large-scale datasets: Reddit, OGB-Proteins, OGB-Products, and IGB. 
    Table~\ref{table:graphsage_large_scale_node_level_task} shows the classification accuracy for Reddit, OGB-Products, and IGB as well as the ROC-AUC for OGB-Proteins, across different $\lambda$ values. The results indicate that MixQ-GNN prediction performance close to FP32 on Reddit and OGB-Proteins, but shows a significant drop on IGB, moderate decline on OGB-Products under aggressive quantization, and achieves an average reduction of $5.6\times$ in BitOPs compared to the FP32 baseline.

    \begin{table}
        \centering
        \resizebox{\linewidth}{!}{
        \begin{tabular}{@{}lccccc@{}}
        \toprule[2pt]
        \shortstack{\textbf{Dataset} \\ \;} & 
        \shortstack{\textbf{$\lambda$ /} \\ \textbf{Precision}} & 
        \shortstack{\textbf{Accuracy /} \\ \textbf{ROC-AUC} $\uparrow$} & 
        \shortstack{\textbf{Bits} \\ \;} & 
        \shortstack{\textbf{GigaBitOps} $\downarrow$  \\ \;} \\
        \midrule[1pt]
        \multirow{4}{*}{\textbf{Reddit}}   & FP32        & $86.72 \pm 0.38\%$          & 32  & 1103 \\
                  & $-10^{-8}$           & $85.50 \pm 1.08\%$          & 6.91 & 129  \\
                  & 0.1                  & $86.01 \pm 1.04\%$          & 5.70 & 111  \\
                  & 1                    & $84.86 \pm 2.38\%$          & 5.21 & 80   \\
        \midrule[1pt]
        \multirow{4}{*}{\textbf{OGB-Proteins}} & FP32 & $0.63$ & 32  & 3369 \\
                  & $10^{-8}$            & $0.61$ & 6.1 & 1299 \\
                  & 0.1                  & $0.61$ & 2.8 & 643  \\
                  & 1.0                    & $0.59$ & 2.4 & 391  \\ 
        \midrule[1pt]
        \multirow{4}{*}{\textbf{OGB-Products}} & FP32 & $66.60 \pm 1.30\%$ & 32  & 1862 \\
                  & $10^{-8}$            & $66.36 \pm 0.98\%$ & 7.5 & 425  \\
                  & 0.1                  & $63.43 \pm 3.93\%$ & 7.2 & 403  \\
                  & 1.0                    & $60.75 \pm 2.24\%$ & 5.0 & 305  \\ 
        \midrule[1pt]
        \multirow{4}{*}{\textbf{IGB}} & FP32 & $71.47 \pm 0.35\%$ & 32  & 14 \\
                  & $10^{-8}$            & $67.25 \pm 1.25\%$ & 6.91 & 1.5  \\
                  & 0.1                  & $67.59 \pm 0.54\%$ & 6.18 & 1.4  \\
                  & 1.0                    & $66.79 \pm 0.30\%$ & 5.45 & 1.2  \\ 
        \bottomrule[2pt]\vspace{0.0em}
        \end{tabular}
        }
        \caption{Native layers of the GraphSAGE architecture with MixQ-GNN perform across four large scale datasets. The evaluation metric for Reddit, Products, and IGB is accuracy, while the Receiver Operating Characteristic - Area Under the Curve (ROC-AUC) is the evaluation metric for Proteins.}
        \label{table:graphsage_large_scale_node_level_task}
    \end{table}

    \subsection{Graph-level Tasks}
    We evaluated MixQ-GNN on various types of graphs from the TUDataset~\cite{Morris2020TuDataset} for graph classification, including bioinformatics and social network graphs. The evaluation used cross-validation across 10 folds. For each fold, we initialized a new relaxed architecture to search for effective bit-width combinations.
    The architecture used was five layers of GIN with MLP of two linear layers, followed by global max pooling across the graph nodes, and then two linear layers with ReLU activation in between to predict the graph classes.
    
    \begin{table}
        \begin{center}
        \renewcommand{\arraystretch}{0.6}
        \resizebox{0.95\linewidth}{!}{
        \begin{tabular}{llccc}
        
        \toprule[2pt]
        \textbf{Dataset}                       & \textbf{Method}   & \textbf{Accuracy $\uparrow$}  & \textbf{Bits} &  \small\textbf{GBitOPs $\downarrow$} \\  
        \midrule[1pt]
        \multirow{7}{*}{\textbf{IMDB-B}}       & FP32                               & 75.2$\pm$4.0\%                        & 32    &   5.47                    \\
                                               & DQ~\cite{tailor2021degreequant}    & 68.6$\pm$7.0\%                        & 4     &   \textcolor{first}{0.68} \\
                                               & DQ~\cite{tailor2021degreequant}    & \textcolor{third}{71.1$\pm$3.9\%}     & 8     &   1.36                    \\
                                               & A$^2$Q~\cite{zhu2023rm}            & \textcolor{first}{74.6$\pm$3.4\%}     & 4.48  &   \textcolor{second}{0.87}\\
        \cmidrule{2-5}
                                               & MixQ $_{(\lambda^\ast)}$           & \textcolor{second}{74.0$\pm$5.6\%}    & 7.83  &   1.27                    \\ 
                                               & MixQ  $_{(\lambda = 1)}$           & 69.6$\pm$7.3\%                        & 5.96  &   \textcolor{third}{1.06} \\ 
        \midrule[1pt]
        \multirow{7}{*}{\textbf{PROTEINS}}     & FP32                               & 70.5$\pm$4.2\%                        & 32    &   7.62                    \\          
                                               & DQ~\cite{tailor2021degreequant}    & \textcolor{second}{73.1$\pm$4.1\%}    & 4     &   \textcolor{first}{0.95} \\
                                               & DQ~\cite{tailor2021degreequant}    & 72.9$\pm$3.5 \%                       & 8     &   \textcolor{third}{1.90} \\
                                               & A$^2$Q~\cite{zhu2023rm}            & \textcolor{first}{74.0$\pm$1.2\%}     & 4.44  &   \textcolor{second}{1.05}\\
        \cmidrule{2-5}
                                               & MixQ $_{(\lambda^\ast)}$           & \textcolor{third}{73.1$\pm$5.5\%}     & 5.81  &   1.35                    \\ 
                                               & MixQ  $_{(\lambda = 1)}$           & 72.8$\pm$5.2\%                        & 5.42  &   1.25                    \\ 
        \midrule[1pt]
        \multirow{7}{*}{\textbf{D\&D}}         & FP32                               & 73.8$\pm$3.3\%                        & 32    &   55.41                   \\
                                               & DQ~\cite{tailor2021degreequant}    & \textcolor{third}{72.7$\pm$2.9\%}     & 4     &   \textcolor{first}{6.92} \\
                                               & DQ~\cite{tailor2021degreequant}    & \textcolor{second}{72.9$\pm$3.1\%}    & 8     &   13.85                   \\
                                               & A$^2$Q~\cite{zhu2023rm}            & 72.2$\pm$1.0\%                        & 4.42  &   10.13                   \\
        \cmidrule{2-5}
                                               & MixQ $_{(\lambda^\ast)}$           & \textcolor{first}{73.7$\pm$6.9\%}     & 4.89  &   \textcolor{second}{8.92}\\ 
                                               & MixQ $_{(\lambda = 1)}$            & 69.6$\pm$10.8\%                       & 4.91  &   \textcolor{third}{9.02} \\ 
        \midrule[1pt]
        \multirow{7}{*}{\textbf{REDDIT-B}}     & FP32                               & 89.54$\pm$1.4\%                       & 32    &   75.68 \\          
                                               & DQ~\cite{tailor2021degreequant}    & 83.4$\pm$4.9\%                        & 4     &   \textcolor{first}{9.46} \\
                                               & DQ~\cite{tailor2021degreequant}    & \textcolor{second}{90.5$\pm$2.0\%}    & 8     &   \textcolor{third}{18.92} \\
                                               & A$^2$Q~\cite{zhu2023rm}            & 88.9$\pm$2.1\%                        & 4.35  &   \textcolor{second}{10.28} \\
        \cmidrule{2-5}
                                               & MixQ $_{(\lambda^\ast)}$           & \textcolor{first}{90.7$\pm$1.5\%}     & 14.97 &   33.63 \\ 
                                               & MixQ  $_{(\lambda = 1)}$           & \textcolor{third}{89.3$\pm$1.5\%}     & 10.32 &   24.34 \\ 
        \midrule[1pt]
        \multirow{7}{*}{\textbf{REDDIT-M}}     & FP32                               & 52.2$\pm$3.2\%     & 32    &   83.70 \\          
                                           & DQ~\cite{tailor2021degreequant}    & 42.7$\pm$2.2\%                        & 4     &   \textcolor{first}{10.46} \\
                                           & DQ~\cite{tailor2021degreequant}    & 50.9$\pm$2.8\%                        & 8     &   \textcolor{third}{20.92} \\
                                           & A$^2$Q~\cite{zhu2023rm}            & \textcolor{first}{54.4$\pm$1.8\%}     & 4.33  &   \textcolor{second}{11.32} \\
        \cmidrule{2-5}
                                           & MixQ $_{(\lambda^\ast)}$           & \textcolor{second}{53.7$\pm$2.4\%}    & 14.77 &   35.62 \\ 
                                           & MixQ  $_{(\lambda = 1)}$           & \textcolor{third}{51.7$\pm$1.9\%}                       & 9.85  &   25.46 \\ 
        \bottomrule[2pt]
        \end{tabular}
        }
        \end{center}
        \caption{Cross-validation accuracy on 10-fold using the GIN architecture. The bit-width search space is $\{4, 8\}$ for the IMDB-B, Proteins, and D\&D datasets, and $\{8, 16\}$ for REDDIT-B,  REDDIT-M datasets. $\lambda^\ast$ indicate the best value for $\lambda$ with respect to the accuracy. The \textcolor{first}{\textbf{first-}}, \textcolor{second}{\textbf{second-}} and \textcolor{third}{\textbf{third-}}best results are colored.}
        \label{table:graph-level-tudataset-dq-vs-a2q-vs-mixq}
    \end{table}    
    
    \begin{table}
        \centering
        \resizebox{0.85\linewidth}{!}{
            \begin{tabular}{llcccc}
                \toprule[2pt]
                \multirow{2}{*}{\textbf{Method}} & \multirow{2}{*}{\textbf{Bits}} & \multicolumn{3}{c}{\textbf{Accuracy $\uparrow$}} \\
                \cline{3-5}
                \cmidrule[0.5pt]{3-5}
                & & \textbf{Mean $\pm$ Std.} & \textbf{Min.} & \textbf{Max.} \\ 
                \midrule[1pt]
                FP32 & 32 & 99.4$\pm$1.3\% & 96.7 & 100.0 \\
                \midrule[0.5pt]
                QAT - INT2 & 2 & 24.4$\pm$8.1\% & 6.7 & 46.7 \\
                QAT - INT4 & 4 & 94.4$\pm$5.9\% & 80.0 & 100.0 \\
                \midrule[0.5pt]
                MixQ\;\:$_{(\lambda=-\varepsilon)}$ & 3.9 & 95.0$\pm$5.1\% & 80.0 & 100.0 \\ 
                MixQ\;\:$_{(\lambda=0)}$ & 3.5 & 94.1$\pm$5.2\% & 76.7 & 100.0 \\ 
                \bottomrule[2pt]\vspace{0.0em}
            \end{tabular}
        }
        \caption{Test results of four GCN layers architecture for CSL dataset.
        Results are from 5-fold cross validation, run 10 times. 
        The dimension of node positional encoding with Laplacian eigenvectors is 50. Epsilon is $\varepsilon = 10^{-3}$.}
        \label{table:csl_mixq_vs_qat}
    \end{table}

    We chose global max pooling to avoid overflow in the quantized values that occur from summation pooling or the floating-point values produced by mean pooling. Max pooling ensures that the quantized values remain within their quantization range.
    In Table~\ref{table:graph-level-tudataset-dq-vs-a2q-vs-mixq}, we reproduced all the results for DQ and A$^2$Q on the TUDataset. As shown, MixQ-GNN maintained accuracy comparable to the full precision architecture. In the case of PROTEINS, REDDIT-B, it achieved higher accuracy with a lower average bit-width and lower bit operations. 

    \subsubsection{Synthetic Dataset}
    We evaluated the prediction performance of quantization on the CSL synthetic dataset~\cite{murphy2019relational} for graph classification. Laplacian Positional Encodings, introduced in~\cite{dwivedi2023benchmarking}, were employed to provide node embeddings.
    In Table~\ref{table:csl_mixq_vs_qat}, we present the test accuracies for native quantization-aware training for INT2, and INT4, and two variation of MixQ-GNN. The table details the mean accuracy, standard deviation, minimum, and maximum values for each method. The four layer GCN architecture starts to achieve reliable accuracy at the level of INT4 quantization. This aligns with the theoretical results from~\cite{aamand2022exponentially}, since the number of nodes in the graphs is $41$, therefore the required number of bits for feature vectors exchanged between nodes is $\log_2(n) = \log_2(41) \approx 5.36$. While MixQ-GNN with less amount of bit archives better accuracy and reduced the number of bits.
    
    On average, MixQ-GNN achieves a bit operation reduction of $5.1\times$ compared to the FP32 architecture across IMDB-B, PROTEINS, D\&D, REDDIT-B, and REDDIT-M datasets.


\section{Ablation Study}
    
    \begin{figure}
        \centering
        \begin{subfigure}[t]{\linewidth}
            \centering
            \includegraphics[width=\textwidth]{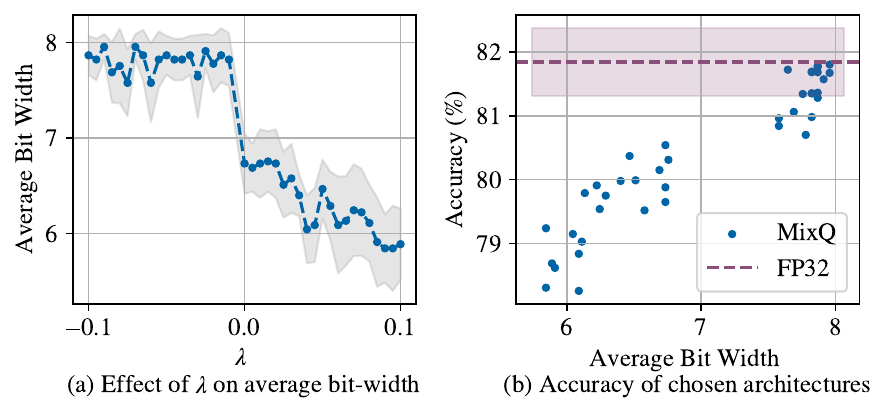}
        \end{subfigure}%
        
        
        \caption{Effect of $\lambda$ on the average bit-width and  accuracy of MixQ-GNN with two layers of the GCN architecture for thirty runs over Cora dataset.}
        \label{figure:lambda_vs_avg_bit_width_and_avg_bit_width_vs_accuracy}
    \end{figure}
    
    In this section, we systematically examine the impact of varying the hyper-parameter $\lambda$ on the average bit-width, and investigate whether MixQ-GNN's gains are substantial or if it merely selects randomly by using random assignment as baselines with a two-layer GCN architecture.    
    
    Figure~\ref{figure:lambda_vs_avg_bit_width_and_avg_bit_width_vs_accuracy}(a) shows experiments with $\lambda$ values from $-0.1$ to $0.1$ and the corresponding average bit-width. For negative $\lambda$ values, the average bit-width varies around $8$ bits. For positive $\lambda$ values, the average bit-width ranges from $5.8$ to $6.7$. Higher $\lambda$ values result in lower average bit-widths, which lead to more computationally efficient architectures.Figure~\ref{figure:lambda_vs_avg_bit_width_and_avg_bit_width_vs_accuracy}(b) presents the corresponding accuracies for each choice from Figure~\ref{figure:lambda_vs_avg_bit_width_and_avg_bit_width_vs_accuracy}(a). Lower bit-widths within the range $[5.8, 6.7]$ slightly reduce accuracy by approximately $2\%$. However, architectures with higher bit-widths within the range $[6.7, 8]$ maintain accuracy levels close to FP32 precision. 
    \subsection{Additive Ablation}
    
    \begin{table}
       \renewcommand{\arraystretch}{1.25}
       \begin{center}
       \resizebox{\linewidth}{!}{
       \begin{tabular}{llccc}
       \toprule[2pt]
       \textbf{Dataset}                       & \textbf{Method}                              & \textbf{Accuracy$\uparrow$}     & \textbf{Bits}        &  \textbf{GBitOPs$\downarrow$} \\  
       \midrule[1pt]
       \multirow{3}{*}{\textbf{Cora}}         & Random                                       & 36.9$\pm$19.5\%       & 4.56${\pm0.7}$         & 2.24${\pm0.6}$     \\ 
                                              & Random + INT8                                & 57.4$\pm$21.4\%       & 4.97${\pm0.6}$         & 2.42${\pm0.7}$     \\ 
                                              & MixQ $_{(\lambda=1)}$                        & \textbf{68.7$\pm$2.7\%}        & 3.84                 & \textbf{1.68}             \\ 
       \midrule[1pt]
       \multirow{3}{*}{\textbf{CiteSeer}}     & Random                                       & 46.1$\pm$15.6\%       & 4.86${\pm0.7}$         & 8.47${\pm2.3}$     \\ 
                                              & Random + INT8                                & 54.2$\pm$14.9\%       & 4.96${\pm0.7}$         & 7.03${\pm1.8}$     \\ 
                                              & MixQ $_{(\lambda=1)}$                        & \textbf{60.9$\pm$8.7\%}        & 3.44                 & \textbf{4.23}             \\ 
       \midrule[1pt]
       \multirow{3}{*}{\textbf{PubMed}}       & Random                                       & 45.5$\pm$21.9\%       & 4.60${\pm0.7}$         & 6.04${\pm2.0}$     \\ 
                                              & Random + INT8                                & 50.8$\pm$21.0\%       & 4.79${\pm0.7}$         & 5.56${\pm1.7}$     \\ 
                                              & MixQ $_{(\lambda=1)}$                        & \textbf{71.0$\pm$1.8\%}        & 4.09                 & \textbf{4.85}             \\ 
       \bottomrule[2pt]
       \end{tabular}
       }
       \end{center}
       \caption{Comparison of thirty runs of random choices of bit-widths $\{2, 4, 8\}$, random choices of bit-width $8$ bits assigned to the prediction output, and MixQ-GNN with $\lambda = 1$.}
       \label{table:node-level-random-vs-mixq}
    \end{table}
    
    We compare the methods \emph{Random\/}, which selects bit-widths randomly for each component, \emph{Random + INT8\/} which does the same but assigns $8$ bits to the last function for high precision in the prediction output, and MixQ-GNN with $\lambda = 1$ across the Cora, CiteSeer, and PubMed datasets in Table~\ref{table:node-level-random-vs-mixq}.
    
    The results demonstrate that random bit-width choices, both with and without the INT8 constraint for the prediction, tend to result in higher average bit-widths and less accuracy compared to MixQ-GNN. 
    
    In the Cora dataset, the random choices achieve $36\%$ accuracy with an average bit-width of $4.5$. 
    In contrast, our MixQ-GNN method achieves a significantly higher accuracy of $68\%$ with a lower average bit-width of $3.8$. This trend is consistent across the CiteSeer and PubMed datasets.

\section{Conclusion}
    In this paper, we introduce MixQ-GNN, a framework that leverages mixed precision quantization to enhance GNN efficiency without sacrificing accuracy. Our framework integrates quantization-aware training into GNNs, enabling the creation of fully quantized MPNNs.
    MixQ-GNN systematically navigates a wide set of possible bit-width combinations within GNN components and ensures effective prediction performance while reducing computational costs.

    Experiments on node and graph classification tasks illustrate that MixQ-GNN maintains or surpasses the accuracy of full precision (FP32) architectures while achieving substantial reductions in computational cost. On average, MixQ-GNN achieved a $5.5\times$ reduction in bit operations for node classification tasks and a $5.1\times$ reduction in bit operations for graph classification tasks compared to full precision  architectures. This highlights its effectiveness in balancing computational efficiency and prediction performance.

    Furthermore, integrating MixQ-GNN with existing quantization methods, such as the DQ quantizer, underscores its versatility and further enhances prediction performance by leveraging both graph structure-aware quantization and optimal bit-width selection. Our results confirm the practical applicability and efficiency of MixQ-GNN in large-scale graph applications.

\section{Acknowledgment}
Nils Kriege was supported by the Vienna Science and Technology Fund (WWTF) [10.47379/VRG19009].

\bibliographystyle{ieeetr} 
\normalsize
\bibliography{references}

\begin{thebibliography}{10}

\bibitem{warden2019tinyml}
P.~Warden and D.~Situnayake, {\em TinyML: Machine Learning with TensorFlow Lite
  on Arduino and Ultra-Low-Power Microcontrollers}.
\newblock O'Reilly Media, 2019.

\bibitem{lin2022ondevice}
J.~Lin, L.~Zhu, W.-M. Chen, W.-C. Wang, C.~Gan, and S.~Han, ``On-device
  training under 256kb memory,'' in {\em Proceedings of the Annual Conference
  on Neural Information Processing Systems (NeurIPS 2022)}, 2022.

\bibitem{Augustin2016ASO}
A.~Augustin, J.~Yi, T.~H. Clausen, and W.~M. Townsley, ``A study of lora: Long
  range and low power networks for the internet of things,'' {\em Sensors},
  2016.

\bibitem{Nagel2021AWP}
M.~Nagel, M.~Fournarakis, R.~A. Amjad, Y.~Bondarenko, M.~v. Baalen, and
  T.~Blankevoort, ``A white paper on neural network quantization,'' {\em arXiv
  preprint arXiv:2103.08499}, 2021.

\bibitem{Zhiqiang2024LLGNN}
Z.~Que, H.~Fan, M.~Loo, H.~Li, M.~Blott, M.~Pierini, A.~Tapper, and W.~Luk,
  ``Ll-gnn: Low-latency graph neural networks on fpgas for high-energy
  physics,'' {\em ACM Transactions on Embedded Computing Systems}, 2024.

\bibitem{Austin2021ETAPrediction}
A.~Derrow-Pinion, J.~She, D.~Wong, O.~Lange, T.~Hester, L.~Perez, M.~Nunkesser,
  S.~Lee, X.~Guo, B.~Wiltshire, P.~W. Battaglia, V.~Gupta, A.~Li, Z.~Xu,
  A.~Sanchez-Gonzalez, Y.~Li, and P.~Velickovic, ``Eta prediction with graph
  neural networks in google maps,'' in {\em Proceedings of the 30th ACM
  International Conference on Information and Knowledge Management (CIKM '21)},
  ACM, 2021.

\bibitem{Weijing2020PointGNN}
W.~Shi and R.~Rajkumar, ``Point-gnn: Graph neural network for 3d object
  detection in a point cloud,'' in {\em Proceedings of the 2020 IEEE/CVF
  Conference on Computer Vision and Pattern Recognition (CVPR)}, 2020.

\bibitem{tailor2021degreequant}
S.~A. Tailor, J.~Fernandez-Marques, and N.~D. Lane, ``Degree-quant:
  Quantization-aware training for graph neural networks,'' in {\em Proceedings
  of the International Conference on Learning Representations}, 2021.

\bibitem{Goodfellow-et-al-2016}
I.~Goodfellow, Y.~Bengio, and A.~Courville, {\em Deep Learning}.
\newblock MIT Press, 2016.

\bibitem{Chen2019MeasuringAR}
D.~Chen, Y.~Lin, W.~Li, P.~Li, J.~Zhou, and X.~Sun, ``Measuring and relieving
  the over-smoothing problem for graph neural networks from the topological
  view,'' in {\em Proceedings of the AAAI Conference on Artificial
  Intelligence}, 2019.

\bibitem{topping2022understanding}
J.~Topping, F.~Di~Giovanni, B.~P. Chamberlain, X.~Dong, and M.~M. Bronstein,
  ``Understanding over-squashing and bottlenecks on graphs via curvature,'' in
  {\em Proceedings of the International Conference on Learning Representations
  (ICLR)}, 2022.

\bibitem{Peng2024BeyondOU}
J.~Peng, R.~Lei, and Z.~Wei, ``Beyond over-smoothing: Uncovering the
  trainability challenges in deep graph neural networks,'' in {\em Proceedings
  of the 33rd ACM International Conference on Information and Knowledge
  Management (CIKM '24)}, ACM, 2024.

\bibitem{Xu2021OptimizationOG}
K.~Xu, M.~Zhang, S.~Jegelka, and K.~Kawaguchi, ``Optimization of graph neural
  networks: Implicit acceleration by skip connections and more depth,'' in {\em
  Proceedings of the 38th International Conference on Machine Learning (ICML
  2021)}, 2021.

\bibitem{Wu2022GraphNN}
L.~Wu, P.~Cui, J.~Pei, L.~Zhao, and X.~Guo, {\em Graph Neural Networks:
  Foundations, Frontiers, and Applications}.
\newblock Springer Singapore, 1st~ed., 2022.

\bibitem{Wentao2021EdgeComputing}
W.~Chen, H.~Qiu, J.~Zhuang, C.~Zhang, Y.~Hu, Q.~Lu, T.~Wang, Y.~Shi, M.~Huang,
  and X.~Xu, ``Quantization of deep neural networks for accurate edge
  computing,'' {\em Journal of Emerging Technologies in Computing Systems},
  2021.

\bibitem{zhu2023rm}
Z.~Zhu, F.~Li, Z.~Mo, Q.~Hu, G.~Li, Z.~Liu, X.~Liang, and J.~Cheng,
  ``Aggregation-aware quantization for graph neural networks,'' in {\em
  Proceedings of the Eleventh International Conference on Learning
  Representations}, 2023.

\bibitem{kipf2017semisupervised}
T.~N. Kipf and M.~Welling, ``Semi-supervised classification with graph
  convolutional networks,'' in {\em Proceedings of the International Conference
  on Learning Representations (ICLR 2017)}, 2017.

\bibitem{velickovic2018graph}
P.~Veličković, G.~Cucurull, A.~Casanova, A.~Romero, P.~Liò, and Y.~Bengio,
  ``Graph attention networks,'' in {\em Proceedings of the International
  Conference on Learning Representations (ICLR 2018)}, 2018.

\bibitem{xu2018how}
K.~Xu, W.~Hu, J.~Leskovec, and S.~Jegelka, ``How powerful are graph neural
  networks?,'' in {\em Proceedings of the International Conference on Learning
  Representations (ICLR 2019)}, 2019.

\bibitem{ijcai2021p214}
Y.~Shi, Z.~Huang, S.~Feng, H.~Zhong, W.~Wang, and Y.~Sun, ``Masked label
  prediction: Unified message passing model for semi-supervised
  classification,'' in {\em Proceedings of the Thirtieth International Joint
  Conference on Artificial Intelligence (IJCAI '21)}, 2021.

\bibitem{Du2017TopologyAG}
J.~Du, S.~Zhang, G.~Wu, J.~M.~F. Moura, and S.~Kar, ``Topology-adaptive graph
  convolutional networks,'' {\em arXiv preprint arXiv:1710.10370}, 2017.

\bibitem{kim2021how}
D.~Kim and A.~Oh, ``How to find your friendly neighborhood: Graph attention
  design with self-supervision,'' in {\em Proceedings of the International
  Conference on Learning Representations (ICLR 2021)}, 2021.

\bibitem{andersch2022nvidia}
M.~Andersch, G.~Palmer, R.~Krashinsky, N.~Stam, V.~Mehta, G.~Brito, and
  S.~Ramaswamy, ``{NVIDIA Hopper Architecture In-Depth},'' 2022.

\bibitem{ladder-osdi24}
L.~Wang, L.~Ma, S.~Cao, Q.~Zhang, J.~Xue, Y.~Shi, N.~Zheng, Z.~Miao, F.~Yang,
  T.~Cao, Y.~Yang, and M.~Yang, ``Ladder: Enabling efficient low-precision deep
  learning computing through hardware-aware tensor transformation,'' in {\em
  Proceedings of the 18th USENIX Symposium on Operating Systems Design and
  Implementation (OSDI '24)}, 2024.

\bibitem{nvidia_blackwell_architecture}
{NVIDIA Corporation}, ``{NVIDIA Blackwell Architecture Technical Brief},''
  2024.

\bibitem{Ozaki2024ExtensionOA}
K.~Ozaki, D.~Mukunoki, and T.~Ogita, ``Extension of accurate numerical
  algorithms for matrix multiplication based on error-free transformation,''
  {\em Japan Journal of Industrial and Applied Mathematics}, 2024.

\bibitem{Gilmer2017MPNN}
J.~Gilmer, S.~S. Schoenholz, P.~F. Riley, O.~Vinyals, and G.~E. Dahl, ``Neural
  message passing for quantum chemistry,'' in {\em Proceedings of the 34th
  International Conference on Machine Learning (ICML 2017)}, 2017.

\bibitem{Hamilton2017InductiveRL}
W.~L. Hamilton, Z.~Ying, and J.~Leskovec, ``Inductive representation learning
  on large graphs,'' in {\em Proceedings of the 31st International Conference
  on Neural Information Processing Systems (NeurIPS 2017)}, 2017.

\bibitem{Bengio2013EstimatingOP}
Y.~Bengio, N.~Léonard, and A.~C. Courville, ``Estimating or propagating
  gradients through stochastic neurons for conditional computation,'' {\em
  arXiv preprint arXiv:1308.3432}, 2013.

\bibitem{Benoit2018Quantization}
B.~Jacob, S.~Kligys, B.~Chen, M.~Zhu, M.~Tang, A.~Howard, H.~Adam, and
  D.~Kalenichenko, ``Quantization and training of neural networks for efficient
  integer-arithmetic-only inference,'' in {\em Proceedings of the IEEE/CVF
  Conference on Computer Vision and Pattern Recognition (CVPR 2018)}, 2018.

\bibitem{Yidi2021VertexCentric}
Y.~Wu, Y.~Gui, T.~Jin, J.~Cheng, X.~Yan, P.~Yin, Y.~Cai, B.~Tang, and F.~Yu,
  ``Vertex-centric visual programming for graph neural networks,'' in {\em
  Proceedings of the 2022 International Conference on Management of Data
  (SIGMOD '22)}, ACM, 2022.

\bibitem{Iqbal2021ReGraphX}
A.~I. Arka, J.~R. Doppa, P.~P. Pande, B.~K. Joardar, and K.~Chakrabarty,
  ``Regraphx: {NoC}-enabled 3d heterogeneous {ReRAM} architecture for training
  graph neural networks,'' in {\em Proceedings of the 2021 Design, Automation
  and Test in Europe Conference and Exhibition (DATE '21)}, 2021.

\bibitem{Zhangxiaowen2022Graphite}
Z.~Gong, H.~Ji, Y.~Yao, C.~W. Fletcher, C.~J. Hughes, and J.~Torrellas,
  ``Graphite: Optimizing graph neural networks on {CPUs} through cooperative
  software-hardware techniques,'' in {\em Proceedings of the 2022 ACM
  International Conference on Architectural Support for Programming Languages
  and Operating Systems (ASPLOS '22)}, ACM, 2022.

\bibitem{Zhiqiang2022Graphiler}
Z.~Xie, M.~Wang, Z.~Ye, Z.~Zhang, and R.~Fan, ``Graphiler: Optimizing graph
  neural networks with message passing data flow graph,'' in {\em Proceedings
  of the 2022 Machine Learning and Systems Conference (MLSys '22)}, 2022.

\bibitem{Xiaobing2022Rubik}
X.~Chen, Y.~Wang, X.~Xie, X.~Hu, A.~Basak, L.~Liang, M.~Yan, L.~Deng, Y.~Ding,
  Z.~Du, and Y.~Xie, ``Rubik: A hierarchical architecture for efficient graph
  neural network training,'' {\em IEEE Transactions on Computer-Aided Design of
  Integrated Circuits and Systems}, 2022.

\bibitem{Chen2021AUL}
T.~Chen, Y.~Sui, X.~Chen, A.~Zhang, and Z.~Wang, ``A unified lottery ticket
  hypothesis for graph neural networks,'' in {\em Proceedings of the 38th
  International Conference on Machine Learning (ICML 2021)}, 2021.

\bibitem{Liu2022ComprehensiveGG}
C.~Liu, X.~Ma, Y.~Zhan, L.~Ding, D.~Tao, B.~Du, W.~Hu, and D.~P. Mandic,
  ``Comprehensive graph gradual pruning for sparse training in graph neural
  networks,'' {\em IEEE Transactions on Neural Networks and Learning Systems},
  2022.

\bibitem{Zeng2020GraphSAINT}
H.~Zeng, H.~Zhou, A.~Srivastava, R.~Kannan, and V.~Prasanna, ``Graph{SAINT}:
  Graph sampling based inductive learning method,'' in {\em Proceedings of the
  International Conference on Learning Representations (ICLR 2020)}, 2020.

\bibitem{Fey2021GNNAutoScaleSA}
M.~Fey, J.~E. Lenssen, F.~Weichert, and J.~Leskovec, ``{GNNAutoScale}: Scalable
  and expressive graph neural networks via historical embeddings,'' in {\em
  Proceedings of the 38th International Conference on Machine Learning (ICML
  2021)}, 2021.

\bibitem{Ding2021VQGNNAU}
M.~Ding, K.~Kong, J.~Li, C.~Zhu, J.~P. Dickerson, F.~Huang, and T.~Goldstein,
  ``Vq-gnn: A universal framework to scale up graph neural networks using
  vector quantization,'' in {\em Proceedings of the 35th International
  Conference on Neural Information Processing Systems (NeurIPS 2021)}, 2021.

\bibitem{Huang2022EPQuantAG}
L.~Huang, Z.~Zhang, Z.~Du, S.~Li, H.~Zheng, Y.~Xie, and N.~Tan, ``Epquant: A
  graph neural network compression approach based on product quantization,''
  {\em Neurocomputing}, 2022.

\bibitem{Feng2020SGQuantST}
B.~Feng, Y.~Wang, X.~Li, S.~Yang, X.~Peng, and Y.~Ding, ``Sgquant: Squeezing
  the last bit on graph neural networks with specialized quantization,'' in
  {\em Proceedings of the 2020 IEEE 32nd International Conference on Tools with
  Artificial Intelligence (ICTAI)}, 2020.

\bibitem{Gao2020GraphNA}
Y.~Gao, H.~Yang, P.~Zhang, C.~Zhou, and Y.~Hu, ``Graph neural architecture
  search,'' in {\em Proceedings of the 29th International Joint Conference on
  Artificial Intelligence (IJCAI '20)}, 2020.

\bibitem{Gao2023GraphNASDA}
Y.~Gao, P.~Zhang, H.~Yang, C.~Zhou, Z.~Tian, Y.~Hu, Z.~Li, and J.~Zhou,
  ``Graphnas++: Distributed architecture search for graph neural networks,''
  {\em IEEE Transactions on Knowledge and Data Engineering}, 2023.

\bibitem{Yang2020DistillingKF}
Y.~Yang, J.~Qiu, M.~Song, D.~Tao, and X.~Wang, ``Distilling knowledge from
  graph convolutional networks,'' in {\em Proceedings of the IEEE/CVF
  Conference on Computer Vision and Pattern Recognition (CVPR 2020)}, 2020.

\bibitem{zhang2021graphless}
S.~Zhang, Y.~Liu, Y.~Sun, and N.~Shah, ``Graph-less neural networks: Teaching
  old {MLPs} new tricks via distillation,'' in {\em Proceedings of the
  International Conference on Learning Representations (ICLR 2022)}, 2022.

\bibitem{aamand2022exponentially}
A.~Aamand, J.~Y. Chen, P.~Indyk, S.~Narayanan, R.~Rubinfeld, N.~Schiefer,
  S.~Silwal, and T.~Wagner, ``Exponentially improving the complexity of
  simulating the weisfeiler-lehman test with graph neural networks,'' in {\em
  Proceedings of the 36th International Conference on Neural Information
  Processing Systems (NeurIPS 2022)}, 2022.

\bibitem{Jing2021MetaAggregatorLT}
Y.~Jing, Y.~Yang, X.~Wang, M.~Song, and D.~Tao, ``Meta-aggregator: Learning to
  aggregate for 1-bit graph neural networks,'' in {\em Proceedings of the 2021
  IEEE/CVF International Conference on Computer Vision (ICCV)}, 2021.

\bibitem{Wang2020BinarizedGN}
H.~Wang, D.~Lian, Y.~Zhang, L.~Qin, X.~He, Y.~Lin, and X.~Lin, ``Binarized
  graph neural network,'' {\em World Wide Web}, 2020.

\bibitem{Bahri2020BinaryGN}
M.~Bahri, G.~Bahl, and S.~Zafeiriou, ``Binary graph neural networks,'' in {\em
  Proceedings of the 2020 IEEE/CVF Conference on Computer Vision and Pattern
  Recognition (CVPR)}, 2020.

\bibitem{Zhu2023MEGAAM}
Z.~Zhu, F.~Li, G.~Li, Z.~Liu, Z.~Mo, Q.~Hu, X.~Liang, and J.~Cheng, ``{MEGA}: A
  memory-efficient {GNN} accelerator exploiting degree-aware mixed-precision
  quantization,'' in {\em Proceedings of the 2024 IEEE International Symposium
  on High-Performance Computer Architecture (HPCA)}, 2024.

\bibitem{liu2018darts}
H.~Liu, K.~Simonyan, and Y.~Yang, ``{DARTS}: Differentiable architecture
  search,'' in {\em Proceedings of the International Conference on Learning
  Representations (ICLR 2019)}, 2019.

\bibitem{Cai2020RethinkingDS}
Z.~Cai and N.~Vasconcelos, ``Rethinking differentiable search for
  mixed-precision neural networks,'' in {\em Proceedings of the 2020 IEEE/CVF
  Conference on Computer Vision and Pattern Recognition (CVPR)}, 2020.

\bibitem{Koryakovskiy2023OneShotMF}
I.~Koryakovskiy, A.~Yakovleva, V.~Buchnev, T.~Isaev, and G.~Odinokikh,
  ``One-shot model for mixed-precision quantization,'' in {\em Proceedings of
  the 2023 IEEE/CVF Conference on Computer Vision and Pattern Recognition
  (CVPR)}, 2023.

\bibitem{Gale2020SparseGK}
T.~Gale, M.~Zaharia, C.~Young, and E.~Elsen, ``Sparse {GPU} kernels for deep
  learning,'' in {\em SC20: International Conference for High Performance
  Computing, Networking, Storage and Analysis}, 2020.

\bibitem{Li2022EfficientQS}
S.~Li, K.~Osawa, and T.~Hoefler, ``Efficient quantized sparse matrix operations
  on tensor cores,'' in {\em Proceedings of the International Conference for
  High Performance Computing, Networking, Storage and Analysis (SC '22)}, 2022.

\bibitem{Wang2021QGTCAQ}
Y.~Wang, B.~Feng, and Y.~Ding, ``{QGTC}: Accelerating quantized graph neural
  networks via {GPU} tensor core,'' in {\em Proceedings of the 27th ACM SIGPLAN
  Symposium on Principles and Practice of Parallel Programming (PPoPP '22)},
  2022.

\bibitem{AnselPyTorch2Faster2024}
J.~Ansel, E.~Yang, H.~He, N.~Gimelshein, A.~Jain, M.~Voznesensky, B.~Bao,
  P.~Bell, D.~Berard, E.~Burovski, G.~Chauhan, A.~Chourdia, W.~Constable,
  A.~Desmaison, Z.~DeVito, E.~Ellison, W.~Feng, J.~Gong, M.~Gschwind, B.~Hirsh,
  S.~Huang, K.~Kalambarkar, L.~Kirsch, M.~Lazos, M.~Lezcano, Y.~Liang,
  J.~Liang, Y.~Lu, C.~Luk, B.~Maher, Y.~Pan, C.~Puhrsch, M.~Reso, M.~Saroufim,
  M.~Y. Siraichi, H.~Suk, M.~Suo, P.~Tillet, E.~Wang, X.~Wang, W.~Wen,
  S.~Zhang, X.~Zhao, K.~Zhou, R.~Zou, A.~Mathews, G.~Chanan, P.~Wu, and
  S.~Chintala, ``Pytorch 2: Faster machine learning through dynamic {Python}
  bytecode transformation and graph compilation,'' in {\em 29th ACM
  International Conference on Architectural Support for Programming Languages
  and Operating Systems (ASPLOS '24)}, ACM, 2024.

\bibitem{SdqHuang22h}
X.~Huang, Z.~Shen, S.~Li, Z.~Liu, X.~Hu, J.~Wicaksana, E.~Xing, and K.-T.
  Cheng, ``{SDQ}: Stochastic differentiable quantization with mixed
  precision,'' in {\em Proceedings of the 39th International Conference on
  Machine Learning (ICML 2022)}, PMLR, 2022.

\bibitem{BayesianBitsBaalen}
M.~van Baalen, C.~Louizos, M.~Nagel, R.~A. Amjad, Y.~Wang, T.~Blankevoort, and
  M.~Welling, ``Bayesian bits: Unifying quantization and pruning,'' in {\em
  Proceedings of the 34th International Conference on Neural Information
  Processing Systems (NeurIPS 2020)}, 2020.

\bibitem{SearchLow8bitYang}
Z.~Yang, Y.~Wang, K.~Han, C.~Xu, C.~Xu, D.~Tao, and C.~Xu, ``Searching for
  low-bit weights in quantized neural networks,'' in {\em Proceedings of the
  34th International Conference on Neural Information Processing Systems
  (NeurIPS 2020)}, 2020.

\bibitem{amd_epyc_9534}
{Advanced Micro Devices, Inc. (AMD)}, ``{AMD EPYC™ 9534 Processor},'' 2024.

\bibitem{ARM_Architecture_Manual}
{Arm Limited}, {\em Arm Architecture Reference Manual for {A}-profile
  Architecture}, 2024.

\bibitem{Bisong2019}
E.~Bisong, {\em Google Colaboratory}.
\newblock Apress, 2019.

\bibitem{Zhilin2016RevisitingGraphEmbeddings}
Z.~Yang, W.~W. Cohen, and R.~Salakhutdinov, ``Revisiting semi-supervised
  learning with graph embeddings,'' in {\em Proceedings of the 33rd
  International Conference on Machine Learning (ICML 2016)}, 2016.

\bibitem{hu2020ogb}
W.~Hu, M.~Fey, M.~Zitnik, Y.~Dong, H.~Ren, B.~Liu, M.~Catasta, and J.~Leskovec,
  ``Open graph benchmark: Datasets for machine learning on graphs,'' in {\em
  Proceedings of the 34th International Conference on Neural Information
  Processing Systems (NeurIPS 2020)}, 2020.

\bibitem{igbdatasets}
A.~Khatua, V.~S. Mailthody, B.~Taleka, T.~Ma, X.~Song, and W.-m. Hwu, ``Igb:
  Addressing the gaps in labeling, features, heterogeneity, and size of public
  graph datasets for deep learning research,'' in {\em Proceedings of the 29th
  ACM SIGKDD Conference on Knowledge Discovery and Data Mining (KDD '23)}, KDD
  '23, ACM, 2023.

\bibitem{murphy2019relational}
R.~Murphy, B.~Srinivasan, V.~Rao, and B.~Ribeiro, ``Relational pooling for
  graph representations,'' in {\em Proceedings of the 36th International
  Conference on Machine Learning (ICML 2019)}, 2019.

\bibitem{Morris2020TuDataset}
C.~Morris, N.~M. Kriege, F.~Bause, K.~Kersting, P.~Mutzel, and M.~Neumann,
  ``{TUDataset}: A collection of benchmark datasets for learning with graphs,''
  in {\em Proceedings of the ICML 2020 Workshop on Graph Representation
  Learning and Beyond (GRL+ 2020)}, 2020.

\bibitem{Wang2023LowbitQF}
S.~Wang, B.~Eravci, R.~Guliyev, and H.~Ferhatosmanoglu, ``Low-bit quantization
  for deep graph neural networks with smoothness-aware message propagation,''
  in {\em Proceedings of the 32nd ACM International Conference on Information
  and Knowledge Management (CIKM '23)}, ACM, 2023.

\bibitem{dwivedi2023benchmarking}
V.~P. Dwivedi, C.~K. Joshi, A.~T. Luu, T.~Laurent, Y.~Bengio, and X.~Bresson,
  ``Benchmarking graph neural networks,'' {\em Journal of Machine Learning
  Research (JMLR)}, 2023.

\end{thebibliography}

\end{document}